\documentclass[10pt,journal,compsoc]{IEEEtran}

\ifCLASSOPTIONcompsoc
  \usepackage[nocompress]{cite}
\else
  \usepackage{cite}
\fi

\usepackage{graphicx}
\usepackage{amsmath}
\usepackage{amssymb}
\usepackage{algorithm}
\usepackage{algorithmic}
\usepackage{tocloft}
\usepackage{multirow}
\usepackage{makecell}
\usepackage{url}
\usepackage[table]{xcolor}
\usepackage{colortbl}
\usepackage{caption}
\usepackage{subcaption}
\usepackage{inconsolata}
\usepackage{float}
\usepackage{overpic}
\usepackage{listings}
\usepackage{booktabs}

\usepackage{etoolbox}
\makeatletter
\AfterEndEnvironment{algorithm}{\let\@algcomment\relax}
\let\@algcomment\relax
\newcommand\algcomment[1]{\def\@algcomment{\footnotesize#1}}
\makeatother

\definecolor{mygray}{gray}{0.87}
\definecolor{ForestGreen}{RGB}{34,139,34}
\definecolor{twiggreen}{RGB}{221,238,214}
\definecolor{twiggreenplus}{RGB}{198,224,185}
\newcommand{\fg}[1]{\mathbf{\mathcolor{ForestGreen}{#1}}}
\definecolor{Forestred}{RGB}{220,50,50}

\definecolor{iccvblue}{RGB}{54,125,189}  % Define iccvblue color
\newcommand\ChangeRT[1]{\noalign{\hrule height #1}}

\begin{document}

\title{Growing a Multi-head Twig via Distillation and Reinforcement Learning to Accelerate Large Vision-Language Models}

\author{Zhenwei~Shao,~
        Mingyang~Wang,~
        Weijun Zhang,~
        Zhou~Yu,~
        Wenwen~Pan,~\\
        Yan~Yang,~
        Tao~Wei,~
        Hongyuan~Zhang,~
        Jun~Yu
\IEEEcompsocitemizethanks{
\IEEEcompsocthanksitem This work was supported in part by the National Natural Science Foundation of China under Grants (62422204, 62125201, U24B20174, 62402152, 62406093), the Key Research and Development Program of Zhejiang Province (No. 2025C01026), in part by the Zhejiang Provincial Natural Science Foundation of China under Grants LRG26F020001, LQN25F020017, LQ24F020032, and in part by the Scientific Research Innovation Capability Support Project for Young Faculty. (Corresponding author: Zhou Yu.)
\IEEEcompsocthanksitem Z. Shao, M. Wang, W. Zhang, Z. Yu, W. Pan, and Y. Yang are with the Zhejiang Key Laboratory of Space Information Sensing and Transmission, School of Computer Science, Hangzhou Dianzi University, China. (e-mail: \{shaozw, kaka\_wangmy, zhangwj, yuz, panww, yangyan\}@hdu.edu.cn)
\IEEEcompsocthanksitem T. Wei and H. Zhang are with Li Auto Inc., China. (e-mail: \{weitao, zhanghongyuan\}@lixiang.com)
\IEEEcompsocthanksitem J. Yu is with the School of Intelligence Science and Engineering, Harbin Institute of Technology (Shenzhen), China. (e-mail: yujun@hit.edu.cn)
\IEEEcompsocthanksitem Work was done when Z. Shao and M. Wang were interns at Li Auto Inc.  
}
}

\markboth{A Submission to IEEE Transactions on Pattern Analysis and Machine Intelligence}%
{Shao \MakeLowercase{\textit{et al.}}: Growing a Multi-head Twig via Distillation and Reinforcement Learning to Accelerate Large Vision-Language Models}

\IEEEtitleabstractindextext{
\begin{abstract}
Large vision-language models (VLMs) have demonstrated remarkable capabilities in open-world multimodal understanding, yet their high computational overheads pose great challenges for practical deployment. Some recent works have proposed methods to accelerate VLMs by pruning redundant visual tokens guided by the attention maps of VLM's early layers. Despite the success of these token pruning methods, they still suffer from two major shortcomings: (i) considerable accuracy drop due to insensitive attention signals in early layers, and (ii) limited speedup when generating long responses (\emph{e.g.}, 30 tokens). To address the limitations above, we present TwigVLM---a simple and general architecture by ``growing'' a lightweight module, named \emph{twig}, upon an early layer of the base VLM. Compared with most existing VLM acceleration methods purely based on visual token pruning, our TwigVLM not only achieves \textbf{better accuracy retention} by employing a twig-guided token pruning (TTP) strategy, but also yields \textbf{higher generation speed} by utilizing a self-speculative decoding (SSD) strategy. Taking LLaVA-1.5-7B as the base VLM, experimental results show that TwigVLM preserves 96\% of the original performance after pruning 88.9\% of visual tokens and achieves 154\% speedup in generating long responses, delivering significantly better performance in terms of both accuracy and speed over the state-of-the-art VLM acceleration methods. Moreover, we extend TwigVLM to an improved TwigVLM++ variant by introducing a novel multi-head twig architecture with a specialized \emph{pruning head}. TwigVLM++ improves pruning quality via a two-stage training paradigm combining a distillation learning stage and a pruning-oriented reinforcement learning stage, and further accelerates inference via a tree-based SSD strategy. TwigVLM++ outperforms TwigVLM by 1.7\% accuracy improvements and 43\% speedup under the same settings, showing both the efficacy and efficiency of the extended method.
\end{abstract}

}

\maketitle

\IEEEdisplaynontitleabstractindextext

\IEEEpeerreviewmaketitle

\IEEEraisesectionheading{\section{Introduction}\label{sec:introduction}}

\IEEEPARstart{T}{he} revolution in large language models (LLMs) has reshaped the landscape of artificial intelligence \cite{gpt3,deepseekr1}. Meanwhile, there has been growing interest in building large vision-language models (VLMs) \cite{gpt4v,llava1,qwenvl,internvl,shao2025imp}, which can perform visual understanding and reasoning and be used in various vision-language tasks such as object grounding \cite{peng2024grounding, ma2024groma}, document understanding \cite{ye2023mplug, luo2024layoutllm}, GUI automation \cite{wang2024mobile, zhang2023appagent}, and robotic control \cite{brohan2023rt, kim2024openvla}. Despite the remarkable progress achieved by these large VLMs, they are usually parametric-intensive and computational-heavy. This poses great challenges for deploying these models in low-latency scenarios. Therefore, the research on VLM acceleration is of practical and emergent demands.

As the computational complexity of transformer architecture \cite{transformer} increases quadratically with token sequence length, reducing the number of tokens offers an effective way to accelerate VLMs. Moreover, visual tokens usually contain much more redundant information than textual tokens \cite{fastv}, especially in high-resolution images and long videos. Therefore, a series of works on visual token pruning has been proposed \cite{fastv, sparsevlm, mustdrop, visionzip}. For instance, FastV~\cite{fastv} leverages attention signals from an early layer of the VLM to drop redundant tokens for subsequent layers. Many following works have extended FastV's idea to progressive token reduction across layers via pruning and merging operations~\cite{sparsevlm,mustdrop}. Meanwhile, some other works have explored using the attention maps from the visual encoder to guide the visual token pruning before being input into the VLM backbone \cite{visionzip,fastervlm}.
Despite the success they have achieved, they still suffer from the following two major shortcomings:

% \vspace{-2pt}
\begin{enumerate}
\renewcommand{\labelenumi}{(\roman{enumi})}
\item To effectively reduce computational costs, a large proportion of visual tokens are usually pruned in the early layers, which is guided by the attention map of the same layer. However, the attention signals in the early layers are insensitive to the task, which restricts the quality of retained visual tokens and leads to a considerable accuracy drop of the accelerated VLM.
\item The VLM inference process consists of the prefilling and decoding stages. Most token pruning-based approaches only focus on the acceleration of the prefilling stage while paying less attention to the decoding stage, which is the most time-consuming part of the inference process. This results in a significant efficiency gap between the theoretical estimation and practical application, especially in the tasks that generate long responses (\emph{e.g.}, more than 30 tokens).
\end{enumerate}

%\vspace{5pt}
We ask: \emph{Is it possible to address the two limitations above in one unified framework?}

\begin{figure}
	\centering
	\includegraphics[width=\columnwidth]{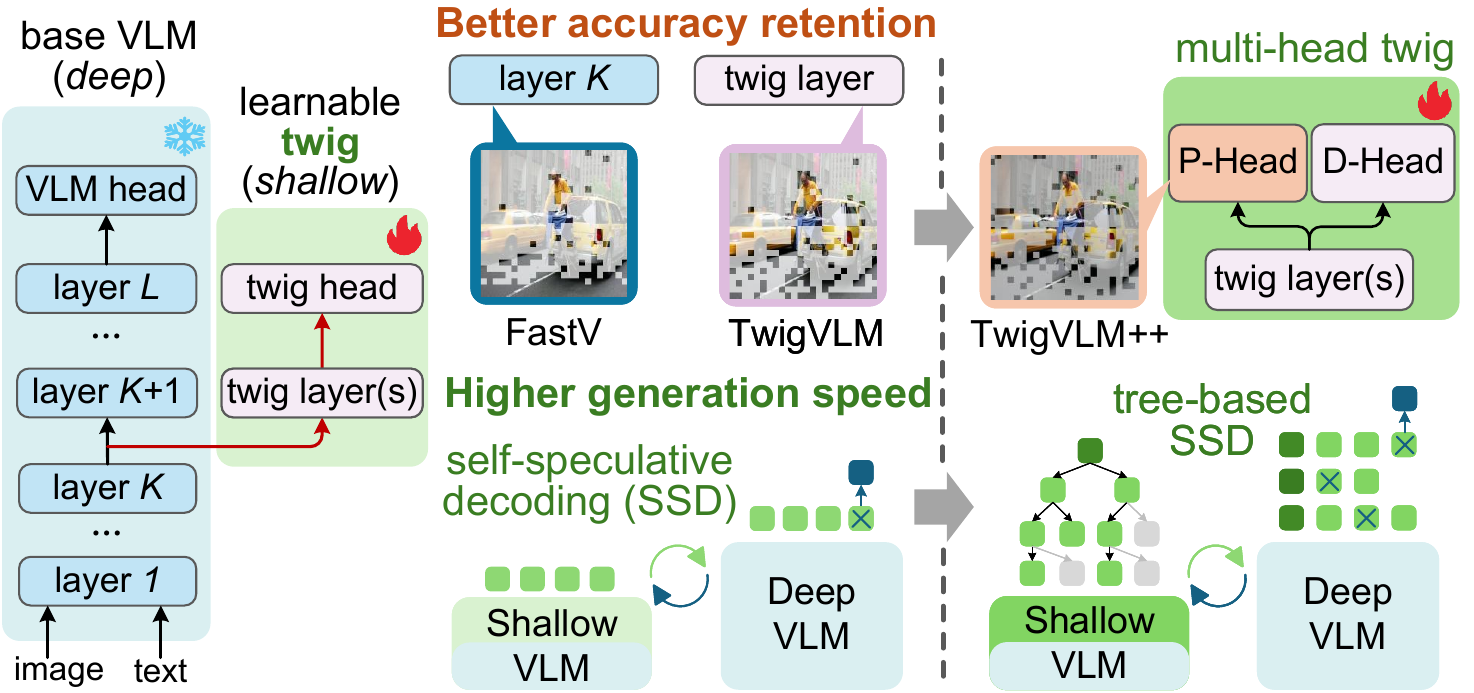}
    \vspace{5pt}
	\begin{minipage}[t]{0.66\linewidth}
	  \centering
	  {\small (a) TwigVLM}
	\end{minipage}
	\begin{minipage}[t]{0.33\linewidth}
	  \raggedright
	  {\small (b) TwigVLM++}
	\end{minipage}
	\vspace{5pt} 
	\includegraphics[width=\columnwidth]{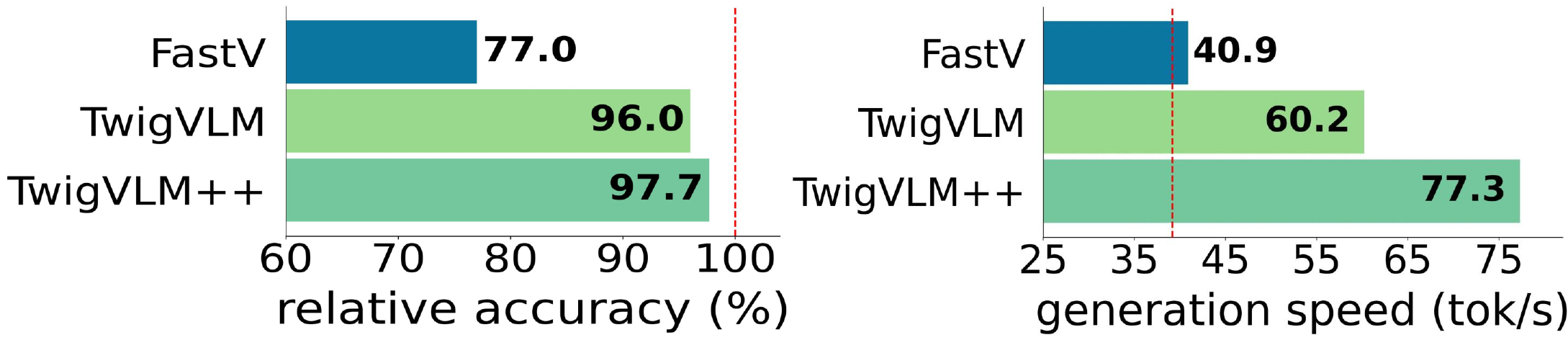} \\
	\vspace{-5pt} 
	\begin{minipage}[t]{1.0\linewidth}
	  \centering
	  {\small (c) Comparisons on accuracy and generation speed.}
	\end{minipage}
	%\vspace{2pt}
    \caption{\textbf{TwigVLM and TwigVLM++ overview}. (a) Given a base deep VLM, the TwigVLM is obtained by freezing the base VLM while training a shallow \emph{twig} block upon its early layer. Compared to the VLM acceleration methods based on visual token pruning (\emph{e.g.}, FastV \cite{fastv}), TwigVLM not only achieves better accuracy retention but also yields higher generation speed. (b) TwigVLM++ extends the design of TwigVLM by further introducing a multi-head twig architecture and a tree-based speculative decoding strategy. (c) The results are evaluated on LLaVA-1.5-7B \cite{llava1.5} (red dotted lines mark its original performance) with the same pruning ratio of 88.9\%.}
	\label{fig:1}
    \vspace{-15pt}
\end{figure}

% \vspace{-5pt}
To this end, we present TwigVLM, a simple and general VLM acceleration approach by appending a lightweight \emph{twig} block upon an early layer of the base VLM. After efficient post-training of the twig block only, the learned twig block is used in both the prefilling and decoding stages of inference. As shown in Fig. \ref{fig:1}a, our TwigVLM not only achieves better accuracy retention by employing a twig-guided token pruning (TTP) strategy, but also yields higher generation speed by utilizing a self-speculative decoding (SSD) strategy \cite{layerskip}. To summarize, TwigVLM is easy to deploy and enables holistic inference acceleration with maximum accuracy retention.

Extensive experiments on a wide range of VLM benchmarks validate the effectiveness of the proposed method. Without bells and whistles, TwigVLM preserves 96\% of the original accuracy after pruning 88.9\% of visual tokens and achieves 154\% speedup in generating long responses. These results demonstrate significant improvements over previous state-of-the-art methods in terms of both accuracy and speed. Moreover, our study also discloses a new aspect for VLM acceleration, \emph{i.e.}, long response generation, which is crucial in real-world scenarios.

A preliminary version of this manuscript was published in \cite{twigvlm_v1}. Based on that version, we have made the following contributions to extend TwigVLM to a more powerful variant TwigVLM++ (see Fig. \ref{fig:1}b): (i) we introduce a novel \emph{multi-head} twig architecture by adding a pruning head (P-Head) alongside the original decoding head (D-Head) in TwigVLM, which decouples the functions of pruning and next-token prediction and leaves design space for pruning-oriented optimization; (ii) we propose a two-stage training paradigm that combines the twig training via distillation learning in the first stage and pruning optimization via reinforcement learning in the second stage; (iii) we adopt a tree-based SSD strategy to replace the original sequence-based SSD, which verifies multiple candidate sequences in parallel to increase the number of accepted tokens per verification step. Extensive and intensive experiments on multiple base VLMs across both image and video benchmarks demonstrate that TwigVLM++ delivers consistent improvements over TwigVLM. As shown in Fig. \ref{fig:1}c, TwigVLM++ outperforms TwigVLM by 1.7\% in accuracy and 43\% in generation speed under the same setting, showing both the efficacy and efficiency of the extended method.

The source code is made available here\footnote{\url{https://github.com/MILVLG/twigvlm}}. We hope our studies may inspire future research on visual token reduction and VLM acceleration.

\section{Related Work}

\noindent\textbf{Inference acceleration for LLMs.}
Accelerating the inference process of LLMs has attracted significant attention from both academia and industry. Although various efficient model architectures have been proposed successively \cite{mamba, yoco, deepseekv3}, they often need to be trained from scratch, thus incurring substantial costs. Consequently, the research of accelerating off-the-shelf LLMs has been extensively investigated from the prefilling and decoding stages of inference, respectively. In the prefilling stage, acceleration is achieved by redundant token reduction  \cite{llmlingua, gist} and attention operator optimization \cite{flashattention, sparseattn}. In the decoding stage, techniques such as dynamic reduction of KV-cache~\cite{streamingllm, h2o} and speculative decoding~\cite{specdec, medusa, liu2025kangaroo, layerskip} are introduced. In particular, speculative decoding performs parallel verification of draft tokens to boost GPU utilization, thus significantly improving the speed.

\vspace{3pt}
\noindent\textbf{Visual token reduction for VLMs.}
Unlike acceleration techniques for LLMs, token reduction methods for VLMs mainly address the problem of visual token redundancy to improve inference efficiency \cite{zhang2024cross}. The pioneering work of FastV \cite{fastv} first studies this redundancy phenomenon, revealing that pruning visual tokens at an early layer of the VLM can achieve acceleration while preserving the majority of the model's accuracy. Following this paradigm, subsequent studies have introduced adaptive or hierarchical pruning strategies \cite{sparsevlm, ye2024fit, mustdrop, ye2024atp, dyvte, FiCoCo-V} and training-based adoption \cite{llavolta,pyramiddrop,yopo,chai2024auroracap} to further reduce the accuracy drop. Meanwhile, there is another line of studies that performs more aggressive token reduction by pruning or merging visual tokens in the visual encoding stage \cite{tome, prumerge,visionzip, fastervlm, fastvlm}. However, the accuracy of these pruned models is suboptimal as they are unaware of the textual prompt when pruning visual tokens. Finally, although the above methods effectively accelerate the prefilling stage of VLMs, their speedup in the decoding stage is limited.

More recently, an emerging line of work has challenged the assumption that attention-based pruning, which tends to retain the most salient tokens, is optimal for minimizing information loss\cite{zhang2025beyond, jiang2025what, endo2025feather, wen2025token}. Instead, these approaches prioritize token representational diversity and global spatial dispersion to reduce redundancy and retain more complementary visual information. 

\vspace{3pt}
\noindent\textbf{Collaborative decoding of small-large VLMs.}
Beyond visual token reduction, some recent works have explored another direction for VLM acceleration, \emph{i.e.}, collaborative decoding between small and large VLMs \cite{gagrani2024speculative,sgl}. Specifically, Gagrani \emph{et al.} introduce a multimodal speculative decoding method, where a small language-only LLM is used to efficiently generate draft tokens and a large VLM then performs verification \cite{gagrani2024speculative}. Similarly, Zhao \emph{et al.} use a small VLM for early-exit prediction, invoking the large VLM only when the prediction confidence is low \cite{sgl}. However, these methods generally rely on a pre-existing yet capable small model with the same vocabulary as the large model, which severely restricts the choice of VLMs. 

To the best of our knowledge, our TwigVLM \cite{twigvlm_v1} is the first work that systematically studies the combination of visual token reduction and speculative decoding within a unified framework. Following this line, several recent works have explored related designs, such as tree-based speculative verification, elastic draft-side visual compression, and verifier-guided video token pruning \cite{huo2025specllava,huang2025specvlmfast,ji-etal-2025-specvlm,kang2025vispec}. Nevertheless, these methods either cannot effectively reduce computation throughout the holistic inference process of VLMs, or exhibit substantially weaker accuracy retention than TwigVLM and its extension TwigVLM++.
\vspace{-5pt}

\section{Pilot Studies}
Before introducing our TwigVLM, we conduct two pilot studies on the accuracy retention and generation speed, respectively, which reveal some crucial but easily overlooked observations in previous VLM acceleration works.
\vspace{-5pt}

\subsection{Preliminaries}
To better understand the pilot studies, we first provide a concise overview of the architecture and two-stage inference process of modern VLMs. After that, we briefly introduce FastV \cite{fastv}, a typical VLM acceleration method via adaptive visual token pruning at inference.

\vspace{3pt}
\noindent\textbf{VLM architecture and inference.} Modern VLMs are usually comprised of a vision encoder and an LLM, where the LLM further consists of a stack of $L$ transformer layers.

During inference, the learned VLM first encodes an input image as a sequence of $M$ visual tokens $\mathbf{X_v}=[\mathbf{v}_1,...,\mathbf{v}_M]\in\mathbb{R}^{M\times d}$ and represents a language prompt as a sequence of $N$ textual tokens $\mathbf{X_q}=[\mathbf{q}_1,...,\mathbf{q}_N]\in\mathbb{R}^{N\times d}$, where $d$ is the common dimensionality of the multimodal tokens. These two groups of tokens are concatenated into $\mathbf{X}=[\mathbf{X_v}, \mathbf{X_q}]$, which is passed through the LLM to generate a sequence of $S$ response tokens $\mathbf{y}=(y_1,...,y_S)$ in an autoregressive manner. At each generation step $j$, we have:
\begin{equation}
y_j=\mathop{\text{argmax}}\limits_{\hat{y}_j} p_\mathrm{LLM}(\hat{y}_{j}|\mathbf{X},\mathbf{y}_{<j}),
\end{equation}
where $\mathbf{y}_{<j}$ is the response sequence before the current token $y_j$.
In practice, the iterative generation process is divided into the \emph{prefilling} and \emph{decoding} stages by leveraging the KV-cache technique \cite{zhou2024survey}.

\vspace{3pt}
\noindent\textbf{Prefilling stage.} This stage executes the forward pass of the input tokens $\mathbf{X}$, which remains unchanged during the response generation. Therefore, their intermediate representations in the LLM, namely the key-value pairs in each self-attention (SA) block of the transformer layer, can be cached and reused in the subsequent decoding stage. Specifically, the SA operation is given by:
\begin{equation}\label{eq:att}
\begin{aligned}
	&\mathrm{SA}(\mathbf{Q},\mathbf{K},\mathbf{V}) = \mathbf{A}\mathbf{V}=\mathrm{softmax}(\mathcal{M}(\mathbf{Q} \mathbf{K}^T/\sqrt{d})) \mathbf{V}\\
	&\mathbf{Q}=\mathbf{X}\mathbf{W_q},\quad\mathbf{K}=\mathbf{X}\mathbf{W_k},\quad \mathbf{V}=\mathbf{X}\mathbf{W_v},
\end{aligned}
\end{equation}
where $\mathcal{M}(\cdot)$ is the causal masking function and $\mathbf{A}$ is the attention map of query-key pairs. $\mathbf{Q},\mathbf{K},\mathbf{V}$ denote the queries, keys, and values, while $\mathbf{W_q}, \mathbf{W_k}, \mathbf{W_v}$ are their linear projection matrices, respectively.

As mentioned, the key-value pairs $\mathbf{K}$ and $\mathbf{V}$ of all transformer layers in the LLM are stored in the KV-cache to expedite the decoding stage.

\vspace{3pt}
\noindent\textbf{Decoding stage.}
The generation of the response token $y_j$ is based on its previous tokens $(\mathbf{X}, \mathbf{y}_{<j})$.
As the attention computation in Eq.(\ref{eq:att}) is causal and parallel for each query, the generation for $y_j$ can be accelerated by feeding only the \emph{single} token into the LLM and computing its corresponding key and value, while reusing the rest key-value pairs from the KV-cache. After each decoding step, the computed key-value pairs for $y_j$ are appended to the KV-cache for the efficient generation of  subsequent tokens.

\begin{figure}
\centering
\includegraphics[width=0.97\linewidth]{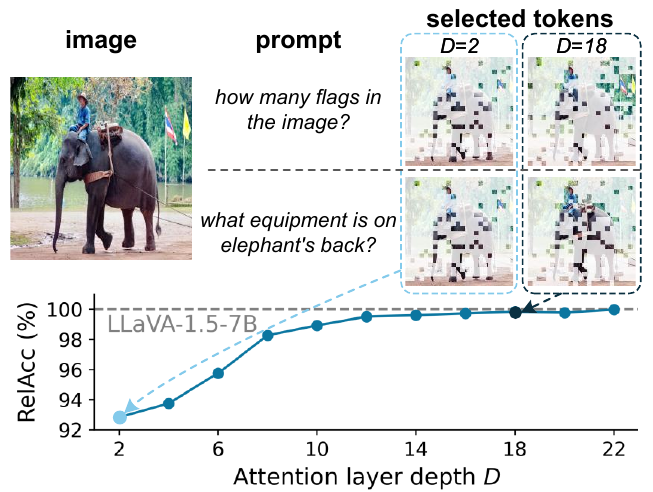}
\caption{\textbf{Attention quality for token selection}. The RelAcc (defined in \S\ref{sec:eval_metric}) is evaluated on GQA \cite{gqa} and TextVQA \cite{textvqa}, which measures the accuracy gap between LLaVA-1.5-7B and its FastV variants with $R$=128 and varied attention layer depth $D$. Taking an image with two prompts as an example, we visualize their attention maps from typical layers $D$=2 and 18, respectively.}
\label{fig:pilot1}
\end{figure}

\vspace{3pt}
\noindent\textbf{Visual token pruning.} The visual tokens in VLMs are much more than the textual tokens and exhibit severe redundancy  \cite{sparsevlm,mustdrop}. To accelerate the inference process of VLMs, some recent works, such as FastV \cite{fastv} and SparseVLM \cite{sparsevlm}, have been proposed to drop redundant visual tokens based on the attention scores during the prefilling stage. Specifically, FastV works by passing the input tokens $\mathbf{X}$ through the first $K$ VLM layers to obtain the attention map $\mathbf{A}^{(K)}$ and token sequence $\mathbf{X}^{(K)}=[\mathbf{X}_\mathbf{v}^{(K)},\mathbf{X}_\mathbf{q}^{(K)}]$ after layer $K$. Denoting $R$ as the number of visual tokens after pruning, the retained token sequence $\mathbf{\hat{X}}^{(K)}$ is obtained as follows:
\begin{equation}\label{eq:token_pruning}
\begin{aligned}
	\mathbf{\hat{X}}^{(K)} &=\mathcal{P}(\mathbf{X}^{(K)}, \mathbf{A}^K, R)\\
	&=\left[ \text{TopR}\left(\mathbf{X}_\mathbf{v}^{(K)}, \sum_{i=M+1}^{M+N} \mathbf{A}^{(K)}[i, 1\text{:}M]
	\right),\mathbf{X}_\mathbf{q}^{(K)} \right].
\end{aligned}
\end{equation}
$\text{TopR}(\mathbf{X}^{(K)}_\mathbf{v}, \mathbf{a})$ selects top-$R$ most significant visual tokens from $\mathbf{X}^{(K)}_\mathbf{v}$, which is guided by the attention scores $\mathbf{a}$ from textual tokens to visual tokens. In FastV, both $K$ and $R$ are set to small values, which means the token lengths are short in most VLM layers. Thus, the computational complexity of the prefilling stage is significantly reduced at the expense of a certain amount of performance drop.
For a fair comparison of different token pruning methods, the average number of retained tokens $\bar{R}$ is used and defined as follows\footnote{The computed result will be rounded to the nearest integer.}:
\begin{equation}\label{eq:avgR}
\bar{R}=[M\times K + R\times (L-K)]/L
\end{equation}

\subsection{Study 1: Attention Quality for Token Selection} \label{sec:pilot_decoding}
The performance of visual token pruning methods heavily depends on the quality of the chosen attention map for pruning. Existing approaches often utilize the attention map from an early layer, \emph{e.g.}, $K$=2, to guide the visual token pruning at the same layer. We ask: \emph{Whether an attention map from a late layer contains more informative signals to facilitate token pruning?} Notably, directly applying the attention map of a late layer to guide the token pruning at an early layer will inevitably introduce redundant computation, so this pilot study only aims to evaluate the quality of the attention maps in different layers.

The pilot experiment is conducted as follows. We use LLaVA-1.5-7B as the base VLM and FastV as the token pruning strategy. Different from the original FastV that selects and prunes visual tokens at the same layer $K$, we perform token selection guided by the attention of the $D$-th layer and then perform token pruning at the $K$-th layer. In this experiment, we set $K$=2 and let $D\geq K$. From the results in Fig. \ref{fig:pilot1}, we can see that the accuracy grows as the attention depth $D$ increases, showing that the attention maps from later layers provide more precise signals for visual token pruning than those from early layers. We hypothesize that later layers are closer to the prediction head (\emph{i.e.}, the loss function), thus their attention maps understand the relation between multimodal tokens more accurately. Moreover, from the visualized attention maps, we find that the selected tokens by early-layer attention ($D$=2) are insensitive to different prompts, while the selected tokens by late-layer attention ($D$=18) are more related to the prompt. This observation also verifies our hypothesis above.

\subsection{Study 2: Prefilling and Decoding Time Costs}\label{sec:pilot_prefilling}
As mentioned above, the inference process consists of the prefilling and decoding stages. FastV and other token-pruning methods primarily accelerate the prefilling stage. We ask: \emph{Can these methods also facilitate the acceleration of the decoding stage, especially when the generated response is relatively long?} This is a practical question since VLMs are often expected to generate long responses (\emph{e.g.}, 30 tokens) in real-world scenarios.

\begin{figure}[t]
\centering
\begin{subfigure}[t]{0.52\columnwidth}
    \centering
    \includegraphics[width=\textwidth]{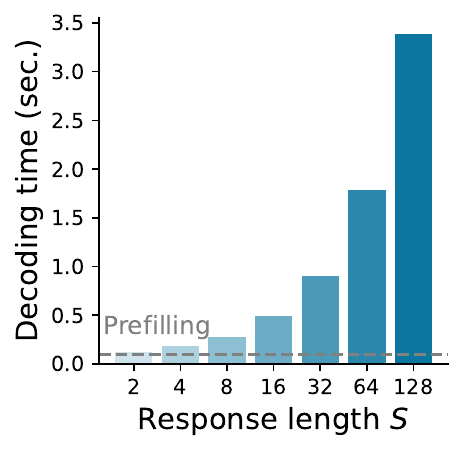}
    \caption{Decoding time with varied $S$}
    \label{fig:pilot2_a}
\end{subfigure}%
\hfill%
\begin{subfigure}[t]{0.48\columnwidth}
    \centering
    \includegraphics[width=\textwidth]{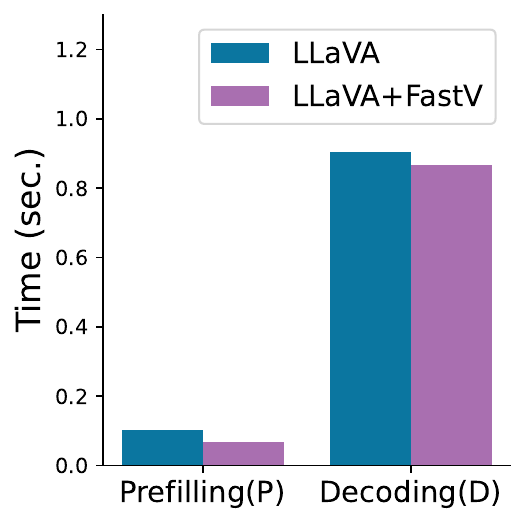}
    \caption{P\&D time at $S$=32}
    \label{fig:pilot2_b}
\end{subfigure}
\caption{\textbf{Prefilling and decoding time costs}. (a) Prefilling time (gray dotted line) and decoding time for LLaVA-1.5-7B \cite{llava1.5} with different response lengths. (b) Prefilling (P) and decoding (D) time comparisons of LLaVA-1.5-7B and its FastV-based variant.}
\label{fig:pilot2}
\end{figure}

\begin{figure*}
\centering
\centering
\begin{subfigure}[b]{0.34\linewidth}
	\centering
	\includegraphics[width=\textwidth]{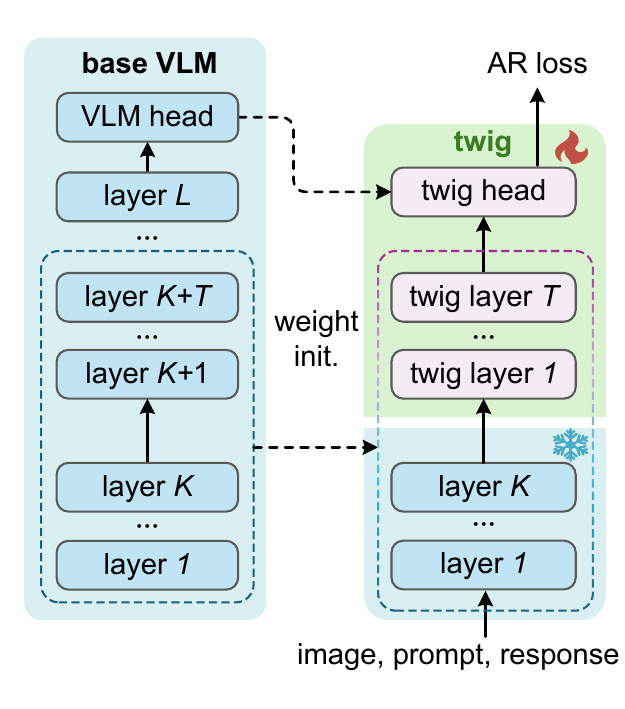}
	\caption{Training stage}
	\label{fig:4a}
\end{subfigure}
\begin{subfigure}[b]{0.33\linewidth}
	\centering
	\includegraphics[width=\textwidth]{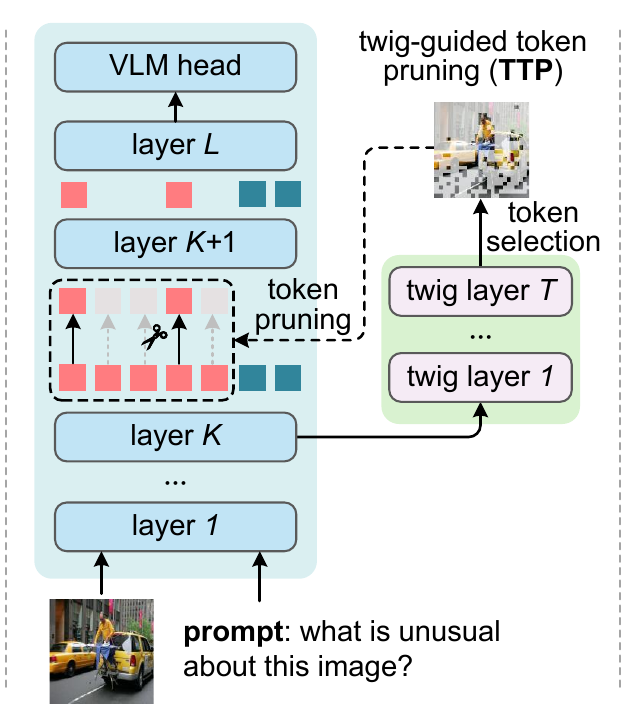}
	\caption{Inference stage (\emph{prefilling})}
	\label{fig:4b}
\end{subfigure}
\begin{subfigure}[b]{0.29\linewidth}
	\centering
	\includegraphics[width=\textwidth]{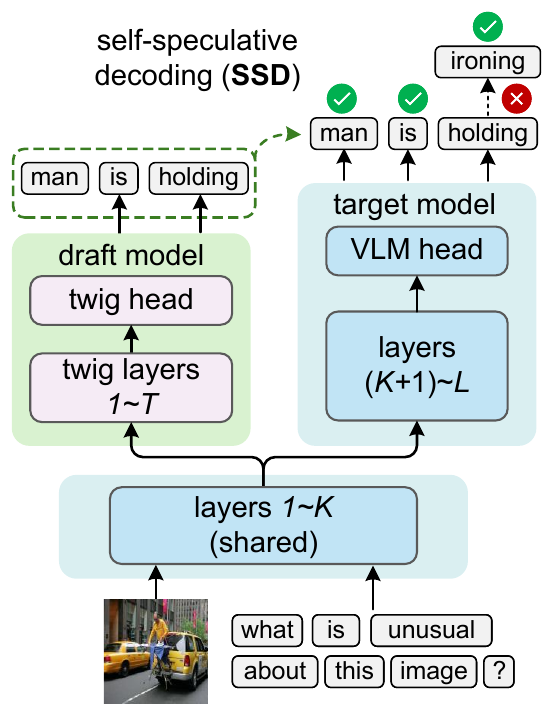}
	\caption{Inference stage (\emph{decoding})}
	\label{fig:4c}
\end{subfigure}
\caption{\textbf{Training and two-stage inference of TwigVLM}. (a) The twig block is initialized from the base VLM and is coupled with the first $K$ layers of base VLM to form a shallow model, which can be trained efficiently. (b) In the prefilling stage, different from previous approaches that perform token selection based on the attention maps from the base VLM, TwigVLM introduces a twig-guided token pruning (TTP) strategy to obtain more precise signals to guide the pruning of visual tokens. (c) In the decoding stage, the shallow and deep sub-networks act as the draft and target models, respectively, enabling it to perform the self-speculative decoding (SSD) \cite{layerskip} to accelerate the generation of long responses.}\label{fig:framework}
\end{figure*}

To answer this question, we conduct the pilot experiment as follows. We take LLaVA-1.5-7B \cite{llava1.5} as the base VLM and compute its time spent on the prefilling and decoding stages for the response length $S$ ranging from 2 to 128. The results in Fig. \ref{fig:pilot2_a} show a linear increase in decoding time w.r.t. the response length $S$, while the prefilling time can be neglected when $S\geq$32.

Next, we compare the prefilling and decoding times of LLaVA-1.5-7B and its FastV-based variant (LLaVA+FastV) when generating responses of the same length $S$=32. From Fig. \ref{fig:pilot2_b}, we observe that although FastV effectively reduces the prefilling time, it attains limited speedup in the decoding stage. This frustrating observation can be explained by two facts: (i) the KV-cache mechanism undermines the speedup for SA blocks achieved by FastV, and (ii) FastV does not accelerate the FFN blocks during decoding, but they account for the majority of the computational costs.

To summarize, these observations highlight the significance and necessity of improving the decoding efficiency for long response generation.

\section{The Proposed TwigVLM}
Inspired by the pilot studies above, we propose TwigVLM, a simple yet effective approach for accelerating VLMs. As depicted in Fig. \ref{fig:framework}, our TwigVLM is obtained by training a lightweight twig block upon an early layer of a frozen VLM. Subsequently, the learned twig block is used in both the prefilling and decoding stages to simultaneously achieve more precise visual token pruning and faster generation speed for long responses.

\subsection{Architecture and Training of TwigVLM}
Denoting a $L$-layer base VLM as $\mathcal{M}_{b}=\{\mathcal{T}_l\}_{l=1}^{L}$, we
attach a twig block $\{\mathcal{G}_t\}_{t=1}^{T}$, \emph{i.e.}, an extra lightweight module with $T$ transformer layer, upon an early layer of $\mathcal{M}_{b}$.
Specifically, the twig block is appended after the $K$-th layer $\mathcal{T}_K$, resembling a twig growing from a tree trunk. When $K+T\ll L$, we obtain a very shallow VLM by combining the twig block and its based $K$ trunk layers $\mathcal{M}_s=\{\mathcal{T}_k\}_{k=1}^{K} \cup \{\mathcal{G}_t\}_{t=1}^{T}$.\footnote{For notational simplicity, we omit the visual encoder and the prediction head in the definition of $\mathcal{M}_b$ and $\mathcal{M}_s$.}

As shown in Fig. \ref{fig:4a}, the training strategy of TwigVLM is simple yet efficient.  All the model weights in the twig block are initialized from the layers $\{\mathcal{T}_i\}_{i=K+1}^{K+T}$ and the prediction head of $\mathcal{M}_{b}$, respectively. After that, the shallow VLM $\mathcal{M}_{s}$ is finetuned using the same training data and the conventional autoregressive (AR) loss as those for $\mathcal{M}_{b}$.
Notably, only the lightweight twig block is updated during the TwigVLM training, consuming only about 10\% of the training time compared to its base VLM.

\subsection{Efficient inference with TwigVLM} \label{sec:inference}
The obtained TwigVLM can be seen as a composition of a deep VLM $\mathcal{M}_b$ and a shallow VLM $\mathcal{M}_s$, where their first $K$ layers are shared with each other. We can leverage this ingenious architecture to accelerate both the prefilling and decoding stages during inference.

\vspace{3pt}
\noindent\textbf{Prefilling acceleration via token pruning.}
The prefilling stage can be accelerated by applying the visual token pruning strategy. Previous methods rely on attention maps from early layers to select key visual tokens and then prune the rest. As discussed in $\S$\ref{sec:pilot_decoding}, we have verified that the attention map of a late layer, which is closer to the prediction head, can provide more precise signals than that of an early layer.
Inspired by this observation, we introduce a twig-guided token pruning (TTP) strategy that utilizes the attention map from the last twig layer to select key tokens.

As shown in Fig. \ref{fig:4b}, given the input tokens $\mathbf{X}$, the TTP strategy first feeds them through the first $K$ layers of $\mathcal{M}_b$ to obtain the latent representations $\mathbf{X}^{(K)}_{\mathcal{M}_b}$, and then feeds them through the twig block to obtain the attention map of the last twig layer $\mathbf{A}^{(K+T)}_{\mathcal{M}_s}$. After that, we use the token pruning function in Eq.(\ref{eq:token_pruning}) to obtain the representations of the retained visual tokens $\mathbf{\hat{X}}^{(K)}_{\mathcal{M}_b}$ as follows:

\begin{equation}\label{eq:pruning}
\mathbf{\hat{X}}^{(K)}_{\mathcal{M}_b} = \mathcal{P}(\mathbf{X}^{(K)}_{\mathcal{M}_b}, \mathbf{A}^{(K+T)}_{\mathcal{M}_s}, R),
\end{equation}
where this condensed token sequence $\mathbf{\hat{X}}^{(K)}_{\mathcal{M}_b}$ is further fed through the remaining layers of $\mathcal{M}_b$.

To achieve aggressive token pruning with small $\bar{R}$, both $K$ and $R$ need to be small values according to Eq.(\ref{eq:avgR}), leading to considerable performance drops. To mitigate this issue, several works adopt \emph{progressive} token pruning strategies~\cite{sparsevlm,pyramiddrop} to retain more tokens at middle layers. In addition, previous studies have shown that the visual tokens in late layers (\emph{e.g.}, after the 20th layer) barely contribute to the prediction~\cite{zhang2024cross,yopo}. Motivated by these works, we additionally introduce a simple \emph{FinalWipe} strategy by removing \emph{all} the visual tokens after the $K_\text{f}$-th layer (a late layer) in $\mathcal{M}_b$. Thus, Eq.(\ref{eq:avgR}) can be reformulated as follows:
\begin{equation}\label{eq:avgR_finalwipe}
\bar{R}=[M\times K + R\times (K_\text{f}-K)]/L.
\end{equation}
When $\bar{R}$ and $K$ are fixed, this strategy enables a larger $R$, which facilitates the performance of our TwigVLM.

\vspace{3pt}
\noindent\textbf{Decoding acceleration via self-speculation.}
The acceleration for the decoding stage remains underexplored, especially for the long response scenarios. TwigVLM's architecture, which contains two sub-networks, namely the deep one $\mathcal{M}_b$ and shallow one $\mathcal{M}_s$, enables the self-speculative decoding (SSD) strategy \cite{layerskip,liu2025kangaroo} to address this challenge. As illustrated in Fig.~\ref{fig:4c}, we take the shallow model $\mathcal{M}_s$ as a \emph{draft} model to efficiently generate multiple subsequent tokens (\emph{i.e.}, draft tokens), and then take the deep model $\mathcal{M}_b$ as a \emph{target} model to verify these tokens in parallel.

Specifically, SSD operates through multiple draft-then-verify iterations to generate the response. In each iteration, the draft model generates multiple tokens in an autoregressive manner. These tokens are then verified by feeding them through the target model in a single forward pass to determine their acceptance. Although the total computational costs are not reduced, the SSD strategy accelerates the decoding stage by processing multiple draft tokens in parallel, which maximally utilizes GPUs' parallel computing capabilities. Moreover, the draft and target models in TwigVLM share the computation and KV-cache of the first $K$ layers, thus achieving further efficiency gains.

It is worth noting that the SSD strategy produces the same response as the target model, which means the accuracy of TwigVLM is only affected by the TTP strategy.

\section{Extending TwigVLM to TwigVLM++}
The pruning capability of the original TwigVLM emerges as a by-product of autoregressive training and is never directly optimized, which limits its effectiveness on stronger or larger base VLMs. To address this, we extend TwigVLM to an improved TwigVLM++ variant by introducing a novel \emph{multi-head} twig architecture with two decoupled heads. Accordingly, we introduce a two-stage training paradigm: (i) twig training via distillation learning (stage-1) and (ii) pruning optimization via reinforcement learning (stage-2). Detailed model architecture and training paradigm are shown in Fig. \ref{fig:10}. Furthermore, we leverage a tree-based self-speculative decoding (SSD) to replace the original sequence-based SSD, yielding higher inference throughput.

\subsection{Multi-head Twig Architecture}
In TwigVLM, the attention map of the last twig layer $\mathcal{G}_T$ serves both autoregressive token prediction and visual token pruning. TwigVLM++ decouples these two functions by extending the twig block with a \emph{multi-head} architecture (see Fig. \ref{fig:1}b), which consists of a \textbf{decoding head} (D-Head) and a \textbf{pruning head} (P-Head). The D-Head retains the standard next-token prediction function of the original twig block. The P-Head is a lightweight auxiliary module attached to the self-attention (SA) layer of $\mathcal{G}_T$, dedicated to computing visual token importance scores.

\vspace{3pt}
\noindent\textbf{P-Head design.}
Let $\mathbf{X}^{(K+T)}$ denote the input to the SA layer of $\mathcal{G}_T$. We extract the hidden state at the last textual token position as $\mathbf{x_q} \in \mathbb{R}^{d}$ and the hidden states at visual token positions as $\mathbf{X_k} \in \mathbb{R}^{M \times d}$. From the projected query and key representations $\mathbf{Q} = \mathbf{X}^{(K+T)}\mathbf{W_q}$ and $\mathbf{K} = \mathbf{X}^{(K+T)}\mathbf{W_k}$, we extract the query vector $\tilde{\mathbf{q}} \in \mathbb{R}^{H \times d_h}$ and key matrix $\tilde{\mathbf{K}} \in \mathbb{R}^{H \times M \times d_h}$, where $H$ is the number of attention heads and $d_h$ is the head dimension. The P-Head applies two learnable gating projections $\mathbf{G}_q$ and $\mathbf{G}_k$ (each a linear layer followed by a nonlinear activation) to modulate these representations. The importance score $\mathbf{s} \in \mathbb{R}^{M}$ is then computed as:
\begin{equation}\label{eq:leaf}
\mathbf{s} = \frac{1}{H}\sum_{h=1}^{H} \sigma\!\left(\frac{\left(\mathbf{G}_q(\mathbf{x_q})^{(h)} \odot \tilde{\mathbf{q}}^{(h)}\right) \left(\mathbf{G}_k(\mathbf{X_k})^{(h)} \odot \tilde{\mathbf{K}}^{(h)}\right)^T}{\sqrt{d_h}}\right),
\end{equation}
where $\sigma(\cdot)$ is the softmax function, $(\cdot)^{(h)}$ denotes the $h$-th head component and $\odot$ is the element-wise product. The resulting $\mathbf{s}$ is a normalized score over visual tokens. During inference, the pruning function in Eq.(\ref{eq:pruning}) is replaced by:
\begin{equation}\label{eq:pruning_leaf}
\mathbf{\hat{X}}^{(K)}_{\mathcal{M}_b} = \mathcal{P}(\mathbf{X}^{(K)}_{\mathcal{M}_b}, \mathbf{s}, R).
\end{equation}
By separating the pruning and prediction pathways, the multi-head architecture provides parameter decoupling between the two tasks, allowing each head to be optimized toward its own objective. More importantly, it creates the design space for targeted pruning optimization, as described in the following training stages.

\begin{figure}
  \centering
  \includegraphics[width=1.0\linewidth]{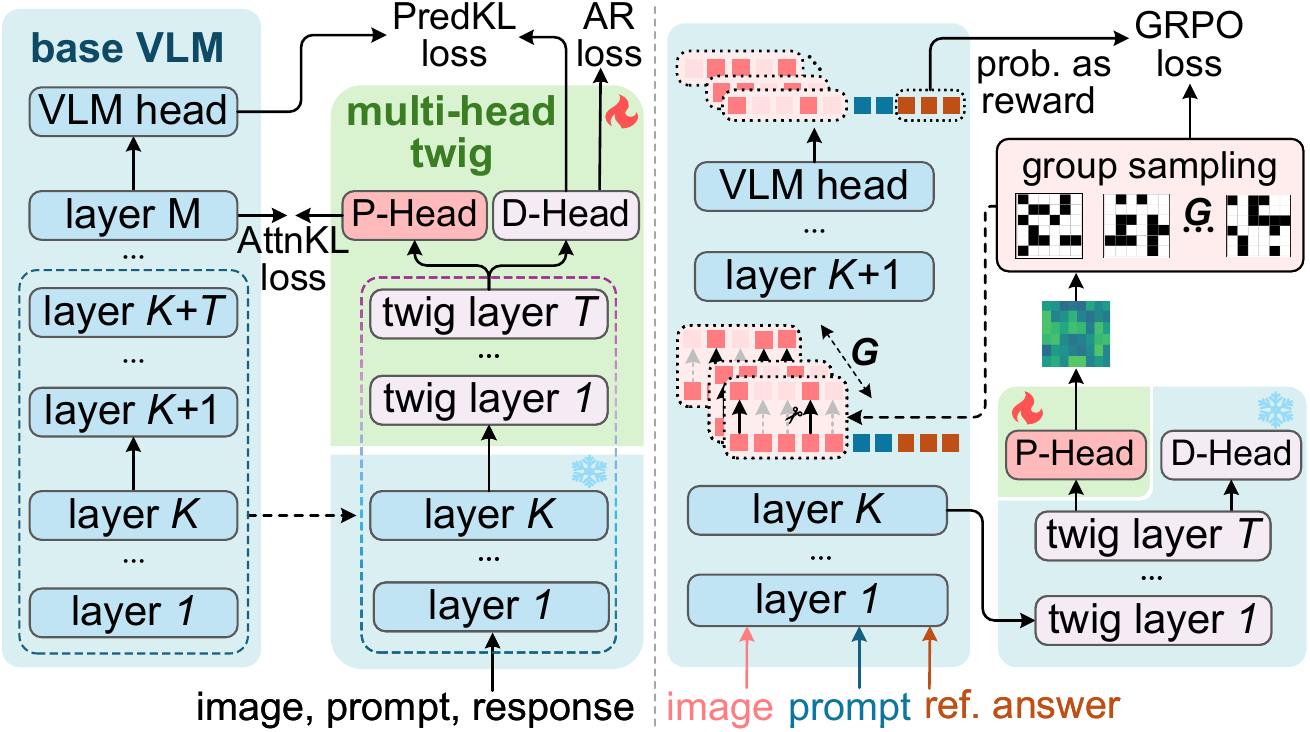}
  %\vspace{-5pt}
  \vspace{-5pt} 
  
  \begin{minipage}[t]{0.5\linewidth}
    \raggedright
    {\small (a) Twig training via distillation learning (\textcolor{ForestGreen}{\S\ref{sec:stage1}})}
  \end{minipage}% 
  \begin{minipage}[t]{0.48\linewidth}
    \raggedright
    {\small (b) Pruning optimization via reinforcement learning (\textcolor{ForestGreen}{\S\ref{sec:stage2}})}
  \end{minipage}
  \vspace{0pt} 
  \caption{The multi-head twig Architecture and two-stage training paradigm of TwigVLM++.}
  \label{fig:twigvlm++}
\end{figure}

\subsection{Twig Training via Distillation Learning} \label{sec:stage1}
As illustrated in Fig. \ref{fig:twigvlm++}(a), in the first training stage, we train the twig block (including the P-Head) using the standard autoregressive next-token prediction (NTP) loss, augmented with two distillation losses that leverage the frozen base VLM $\mathcal{M}_b$ as the teacher.

\vspace{3pt}
\noindent\textbf{PredKL.}
To improve the alignment between the twig (draft) model and the base (target) model, we introduce a prediction-level KL divergence loss:
\begin{equation}\label{eq:predkl}
\mathcal{L}_\text{PredKL} = \mathrm{KL}\!\left(p_{\mathcal{M}_b} \,\|\, p_{\mathcal{M}_s}\right),
\end{equation}
where $p_{\mathcal{M}_s}$ and $p_{\mathcal{M}_b}$ are the next-token prediction distributions of $\mathcal{M}_s$ and $\mathcal{M}_b$, respectively. This strong-to-weak distillation provides the twig block with richer supervisory signals, enhancing its understanding of visual tokens and thereby improving the quality of importance score estimation for pruning.

\vspace{3pt}
\noindent\textbf{AttnKL.}
To directly supervise the pruning signal, we introduce an attention-level KL divergence loss. Let $\mathbf{a}_b \in \mathbb{R}^{M}$ denote the attention distribution from textual tokens to visual tokens at a designated layer of $\mathcal{M}_b$, averaged over all attention heads. The AttnKL loss is defined as:
\begin{equation}\label{eq:attnkl}
\mathcal{L}_\text{AttnKL} = \mathrm{KL}\!\left(\mathbf{a}_b \,\|\, \mathbf{s}\right),
\end{equation}
where $\mathbf{s}$ is the P-Head score from Eq.(\ref{eq:leaf}). This loss guides the P-Head to produce importance scores that are consistent with the attention patterns of the deep model, which has been shown to provide more precise signals for token selection (see $\S$\ref{sec:pilot_decoding}).

The overall loss for the first training stage is:
\begin{equation}\label{eq:stage1_loss}
\mathcal{L}_\text{stage-1} = \mathcal{L}_\text{NTP} + \alpha \cdot \mathcal{L}_\text{PredKL} + \gamma \cdot \mathcal{L}_\text{AttnKL},
\end{equation}
where $\alpha$ and $\gamma$ are two balancing hyper-parameters.

\subsection{Pruning Optimization via Reinforcement Learning} \label{sec:stage2}
After the first stage, the P-Head has acquired a reasonable pruning ability. As illustrated in Fig. \ref{fig:twigvlm++}(b), in the second stage, we further optimize the P-Head's parameters via reinforcement learning (RL) to directly maximize the post-pruning model performance, while the rest of the model weights are frozen. This training stage re-uses the SFT dataset from the first stage but only needs $\sim$10\% of the training samples. Therefore, this training stage is highly efficient in both computation and data.

\vspace{3pt}
\noindent\textbf{Action space.}
We formulate the visual token pruning as a sequential decision process. Given the importance score distribution $\pi(\cdot) = \mathbf{s}$ produced by the P-Head, a pruning action $\mathbf{a}$ consists of selecting $R$ visual token positions to retain. To make this combinatorial selection amenable to policy gradient optimization, we model it as a sequential sampling process without replacement: at each step, a token position is sampled according to the current distribution, then its probability is set to zero, and the distribution is renormalized before the next sampling step. The probability of a complete action $\mathbf{a} = (a_1, \ldots, a_R)$ is:
\begin{equation}\label{eq:action_prob}
\pi_\theta(\mathbf{a}) = \prod_{j=1}^{R} \frac{\pi(a_j)}{\sum_{m \notin \{a_1,\ldots,a_{j-1}\}} \pi(m)}.
\end{equation}

\vspace{3pt}
\noindent\textbf{Reward function.}
Instead of relying on active roll-out and explicit answer verification, we adopt a reference-based reward for RL training \cite{jepo,verifree}. The reward is derived from the model's own log-likelihood on the reference answer---the annotated response readily available in standard SFT data---under the pruned input. Specifically, given a training sample with input $\mathbf{X}$ and ground-truth response $\mathbf{y}^*=(y^*_1,\ldots,y^*_S)$, we apply the pruning action $\mathbf{a}$ to obtain the pruned input $\mathbf{\hat{X}}$, then feed it through the base model $\mathcal{M}_b$ to compute the per-token log-probability of generating $\mathbf{y}^*$. The reward of action $\mathbf{a}$ is defined as:
\begin{equation}\label{eq:reward}
r(\mathbf{a}) = \frac{1}{S}\sum_{j=1}^{S} \log p_{\mathcal{M}_b}(y^*_j \mid \mathbf{\hat{X}}, \mathbf{y}^*_{<j}),
\end{equation}
which is the mean log-probability (i.e., the negative length-normalized cross-entropy loss) of the reference answer. A higher reward indicates that the pruning better preserves the task-relevant information needed for generating the correct response.

\vspace{3pt}
\noindent\textbf{Training procedure.}
We adopt a GRPO-style \cite{grpo} optimization. For each training sample, we sample $G$ pruning actions from the current policy $\pi_\theta$ and compute their rewards. The advantage is estimated via group-level normalization:
\begin{equation}\label{eq:advantage}
\hat{A}_i = \frac{r_i - \text{mean}(\{r_1,r_2,\ldots,r_G\})}{\text{std}(\{r_1,r_2,\ldots,r_G\})},
\end{equation}
where $r_i = r(\mathbf{a}_i)$ are the rewards of the $i$-th action, $\text{mean}(\cdot)$ and $\text{std}(\cdot)$ are the mean and standard deviation functions. Since we perform strict on-policy updates (one policy sampling followed by one gradient step), the importance ratio $\pi_\theta / \pi_{\theta_\text{old}}$ equals 1 and the clipping mechanism in GRPO does not activate. Thus, the loss simplifies to:
\begin{equation}\label{eq:grpo_loss}
\mathcal{L}_\text{stage-2} = \frac{1}{G}\sum_{i=1}^{G} \hat{A}_i \cdot \log \pi_\theta(\mathbf{a}_i).
\end{equation}

\vspace{3pt}
\noindent\textbf{Dynamic pruning-ratio schedule.}
To enable a single trained model to support different pruning ratios at test time without retraining, we randomize $\bar{R}$ during RL training. At each training step, $\bar{R}$ is sampled from a predefined candidate set $\mathcal{R}=\{\bar{R}_1, \ldots, \bar{R}_n\}$ (sorted in ascending order), so that the P-Head learns to produce effective importance scores across a range of pruning ratios. 

Since smaller $\bar{R}$ values (more aggressive pruning) are inherently harder to optimize, we adopt a curriculum-based schedule whose sampling distribution gradually shifts toward smaller values as training progresses:
\begin{equation}\label{eq:elastic}
P(\bar{R}\!=\!\bar{R}_i) = \frac{\exp\!\big({-\beta(t) \cdot i}\big)}{\sum_{j=1}^{n} \exp\!\big({-\beta(t) \cdot j}\big)},\quad i=1,2,\ldots,n,
\end{equation}
where $\beta(t) = \beta_{\max} \cdot ({t}/{T})^{p}$ is an annealing parameter, $t$ and $T$ are the current and total training steps, and $p$ controls the annealing speed. The distribution is uniform at $t\!=\!0$ and concentrates on $\bar{R}_1$ as $t \!\to\! T$, progressively directing the optimization toward the most aggressive pruning ratio.

\subsection{Inference Acceleration via Tree-based SSD}
As described in $\S$\ref{sec:inference}, TwigVLM accelerates the decoding stage via self-speculative decoding (SSD), where $\mathcal{M}_s$ drafts a single token sequence and $\mathcal{M}_b$ verifies it. The efficiency of this approach is limited by the number of accepted tokens per verification step. To improve this, TwigVLM++ adopts a tree-based SSD strategy \cite{specinfer}.

Instead of generating a single draft sequence, $\mathcal{M}_s$ constructs a \emph{token tree} governed by an expansion width $\mathcal{E}$, a selection width $\mathcal{K}$, and a tree depth $\mathcal{D}$. At each level, the top-$\mathcal{K}$ nodes are selected for expansion, and each selected node branches into $\mathcal{E}$ children corresponding to its top-$\mathcal{E}$ predicted tokens from $\mathcal{M}_s$. Each root-to-leaf path in the resulting tree forms a distinct candidate sequence, increasing the coverage of the target model's prediction. For parallel verification, $\mathcal{M}_b$ processes the entire tree in a single forward pass using \emph{tree attention} \cite{specinfer}, which replaces the standard causal mask with a \emph{topology-aware causal mask} so that each node attends only to its ancestors. The verified tokens are determined by traversing the tree from the root until no child matches the target model's output.

Although the drafting and verification of the token tree introduce more computational overheads, tree-based SSD yields more accepted tokens per verification step due to the expanded path coverage, translating to higher decoding throughput in practice.
\vspace{-10pt}

\section{Experiments}
We evaluate the performance of TwigVLM and TwigVLM++ on three popular VLMs, namely LLaVA-1.5-7B \cite{llava1.5}, LLaVA-NeXT-7B \cite{llavanext} and Qwen2.5-VL-7B \cite{qwenvl}, to compare with the state-of-the-art VLM acceleration methods. After that, we conduct comprehensive ablation studies to analyze the effectiveness of the key elements of TwigVLM and TwigVLM++.
\vspace{-10pt}

\subsection{Implementation Details}
Unless otherwise noted, we implement TwigVLM with the following hyper-parameters: the number of twig layers $T$=3, the token pruning position $K$=2, and the FinalWipe position $K_\text{f}$=24.
Based on the default setting above, we adjust $R$ using Eq.(\ref{eq:avgR_finalwipe}) to fairly compare TwigVLM with other methods at the same pruning ratio $1-\bar{R}/M$.

For TwigVLM++, the stage-1 training follows the same protocol as TwigVLM but additionally trains the multiple heads with the distillation losses ($\alpha$=$0.1$, $\gamma$=$1.0$ in Eq.(\ref{eq:stage1_loss})). In stage-2, we train the P-Head using $G$=$32$ groups per sample on $50$K samples from the same SFT dataset as stage-1. The candidate set for the dynamic pruning-ratio schedule is $\mathcal{R}$=\{$64$, $85$, $107$, $128$, $149$, $171$, $192$\}. The annealing parameters are set to $\beta_{\max}$=$8.0$ and $p$=$2.0$. For tree-based SSD, we use the expansion configuration of $\mathcal{E}$=$10$, $\mathcal{K}$=$10$ and $\mathcal{D}$=$4$.

For training data, we use the LLaVA-665K dataset \cite{llava1.5} to train TwigVLM/TwigVLM++ for the LLaVA-1.5-7B and LLaVA-NeXT-7B models, and a dataset of 5M single-image samples from the MAmmoTH-VL-10M dataset \cite{mammoth-VL} for the Qwen2.5-VL-7B model. All experiments are conducted on a server with 8$\times$A100 GPUs. More implementation details are provided in the supplementary.

\begin{table}[t]
\begin{center}
	\setlength{\tabcolsep}{1.8pt}{
		\renewcommand{\arraystretch}{1.3}
		\footnotesize
		\centering
		\begin{tabular}{c | c c c c c c | c }
			\textbf{Method} & \textbf{GQA} & \textbf{MMB} & \textbf{MME} &  \textbf{VQA}$^{\text{T}}$ & \textbf{SQA}$^{\text{I}}$ & \textbf{VQA}$^{\text{V2}}$ & \textbf{RelAcc}\\
			\ChangeRT{1pt}
			\rowcolor{mygray}
			\multicolumn{8}{c}{\textit{Upper Bound, 576 Tokens} \ $\textbf{(100\%)}$} \\
			\textcolor{gray}{~~LLaVA-1.5-7B~~} & \textcolor{gray}{61.9} & \textcolor{gray}{64.7} & \textcolor{gray}{1862} & \textcolor{gray}{58.2} & \textcolor{gray}{69.5} & \textcolor{gray}{78.5} & \textcolor{gray}{100\%}  \\
			\hline
			\rowcolor{mygray}
			\multicolumn{8}{c}{\textit{Retain Averaged 192 Tokens} \ $\fg{(\downarrow 66.7\%)}$}  \\
			FastV~\cite{fastv}                   & 56.5 & 63.7 & 1786 & 57.3 & \textbf{69.5} & 74.6 & 96.5\% \\
			SparseVLM~\cite{sparsevlm}               & 57.6 & 62.5 & 1721 & 56.1 & 69.1 & 75.6 & 95.7\% \\
			PDrop~\cite{pyramiddrop}               & 57.3 & 63.3 & 1797 & 56.5 & \underline{69.2} & 75.1 & 96.5\% \\
			MustDrop~\cite{mustdrop}                & 58.2 & 62.3 & 1787 & 56.5 & \underline{69.2} & 76.0 & 96.6\% \\
			VisionZip~\cite{visionzip}               & 59.3 & 63.0 & 1783 & 57.3 & 68.9 & 76.8 & 97.4\% \\
			VisionZip\ddag~\cite{visionzip}   & 60.1 & 63.4 & 1834 & 57.8 & 68.2 & 77.4 & 98.3\% \\
			\textbf{TwigVLM} &  \underline{61.2} & \underline{64.0} & \underline{1848} & \underline{58.0} & 68.8 & \underline{78.1} & \underline{99.2\%} \\
			\textbf{TwigVLM++} &  \textbf{61.2} & \textbf{64.3} & \textbf{1868} & \textbf{58.0} & \underline{69.2} & \textbf{78.2} & \textbf{99.6\%} \\
			\hline
			\rowcolor{mygray}
			\multicolumn{8}{c}{\textit{Retain Averaged 128 Tokens} \ $\fg{(\downarrow 77.8\%)}$} \\
			FastV                   & 53.0 & 61.4 & 1646 & 56.0 & \underline{69.5} & 69.2 & 92.2\% \\
			SparseVLM               & 56.0 & 60.0 & 1696 & 54.9 & 67.1 & 73.8 & 93.2\% \\
			PDrop                   & 57.1 & 61.6 & 1761 & 56.6 & 68.4 & 72.9 & 95.1\% \\
			MustDrop                & 56.9 & 61.1 & 1745 & 56.3 & 68.5 & 74.6 & 95.1\% \\
			VisionZip               & 57.6 & 62.0 & 1762 & 56.8 & 68.9 & 75.6 & 96.1\% \\
			VisionZip\ddag   & 58.9 & 62.6 & \underline{1823} & 57.0 & 68.3 & 76.6 & 97.3\% \\
			\textbf{TwigVLM} &  \underline{60.6} & \underline{63.5} & 1818 & \underline{57.8} & \underline{69.5} & \underline{77.9} & \underline{98.7\%} \\
			\textbf{TwigVLM++} &  \textbf{60.8} & \textbf{63.7} & \textbf{1856} & \textbf{58.0} & \textbf{69.5} & \textbf{77.9} & \textbf{99.2\%} \\
			\hline
			\rowcolor{mygray}
			\multicolumn{8}{c}{\textit{Retain Averaged 64 Tokens} \ $\fg{(\downarrow 88.9\%)}$}\\
			FastV                   & 44.1 & 45.9 & 1218 & 50.7 & \underline{70.0} & 52.0 & 77.0\% \\
			SparseVLM               & 52.7 & 56.2 & 1505 & 51.8 & 62.2 & 68.2 & 89.9\% \\
			PDrop                   & 47.5 & 58.8 & 1561 & 50.6 & 69.0 & 69.2 & 87.6\% \\
			FasterVLM~\cite{fastervlm}              & 51.5 & 58.5 & 1573 & 53.1 & 69.6 & 66.8 & 89.1\% \\
			MustDrop                & 53.1 & 60.0 & 1612 & 54.2 & 63.4 & 69.3 & 89.6\% \\
			VisionZip               & 55.1 & 60.1 & 1690 & 55.5 & 69.0 & 72.4 & 93.3\% \\
			VisionZip\ddag   & 57.0 & \underline{61.5} & 1756 & \underline{56.0} & 68.8 & 74.2 & 95.2\% \\
			\textbf{TwigVLM} &  \underline{58.8} & 60.4 & \underline{1760} & 55.8 & \textbf{70.0} & \underline{75.6} & \underline{96.0\%} \\
			\textbf{TwigVLM++} &  \textbf{59.7} & \textbf{63.2} & \textbf{1801} & \textbf{56.7} & 69.5 & \textbf{76.8} & \textbf{97.7\%} \\
		\end{tabular}
		\caption{Performance comparisons with three \textcolor[RGB]{34,139,34}{pruning ratios} on six VLM benchmarks \cite{gqa,liu2023mmbench,fu2023mme,textvqa,sciqa,vqav2}. All methods are applied on the same base model \textbf{LLaVA-1.5-7B}. The best result for each benchmark and pruning ratio is \textbf{bolded} and the second best result is \underline{underlined}.}
		\label{tab:llava1.5}
        \vspace{-10pt}
	}
\end{center}
\end{table}

\begin{table}[t]
\centering
\setlength{\tabcolsep}{1.8pt}
\renewcommand{\arraystretch}{1.3}
\footnotesize
	\begin{tabular}{c | c c c c c c | c }
		\textbf{Method} & \textbf{GQA} & \textbf{MMB} & \textbf{MME} & \textbf{VQA}$^{\text{T}}$ & \textbf{SQA}$^{\text{I}}$ & \textbf{VQA}$^{\text{V2}}$ &  \makecell[c]{\textbf{RelAcc}} \\
		\ChangeRT{1pt}
		\rowcolor{mygray}
		\multicolumn{8}{c}{\textit{Upper Bound, 2880 Tokens} \ $\textbf{(100\%)}$} \\
		\textcolor{gray}{LLaVA-NeXT-7B} & \textcolor{gray}{64.2} & \textcolor{gray}{67.9} & \textcolor{gray}{1851} & \textcolor{gray}{61.3}  & \textcolor{gray}{70.2} & \textcolor{gray}{81.8} &  \textcolor{gray}{100\%} \\
		\hline
		\rowcolor{mygray}
		\multicolumn{8}{c}{\textit{Retain Averaged 640 Tokens} \ $\fg{(\downarrow 77.8\%)}$} \\
		FastV       & 62.0 & 65.8 & 1807 & 60.0 & 69.1 & 79.5  & 97.6\% \\
		SparseVLM       & 60.3 & 65.7 & 1772 & 57.8 & 67.7 & 77.1  & 95.4\% \\
		VisionZip       & 61.3 & 66.3 & 1787 & \underline{60.2} & 68.1 & 79.1  & 97.1\% \\
		VisionZip\ddag & 62.4 & 65.9 & 1778 & \textbf{60.8} & 67.9 & 79.9  & 97.4\% \\
		\textbf{TwigVLM} & \underline{63.4} & \underline{67.4} & \textbf{1864} & 58.6 & \underline{69.9} & \underline{81.2} & \underline{99.0\%} \\
		\textbf{TwigVLM++} & \textbf{63.4} & \textbf{67.6} & \underline{1858} & 58.6 & \textbf{70.2} & \textbf{81.2} & \textbf{99.0\%} \\
		\hline
		\rowcolor{mygray}
		\multicolumn{8}{c}{\textit{Retain Averaged 320 Tokens} \ $\fg{(\downarrow 88.9\%)}$} \\
		FastV       & 54.9 & 60.0 & 1539 & 54.8 & 68.2 & 69.6  & 88.2\% \\
		SparseVLM       & 57.7 & 64.3 & 1694 & 55.9 & 67.3 & 73.4  & 92.3\% \\
		MustDrop        & 57.3 & 62.8 & 1641 & \textbf{59.9} & 68.0 & 73.7  & 92.6\% \\
		VisionZip       & 59.3 & 63.1 & 1702 & 58.9 & 67.3 & 76.2  & 93.8\% \\
		VisionZip\ddag & 61.0 & 64.4 & \underline{1770} & \underline{59.3} & 67.5 & 78.4  & 95.8\% \\
		\textbf{TwigVLM}  & \underline{62.2} & \underline{65.0} & 1758 & 57.4 & \underline{68.7} & \underline{79.7}  & \underline{96.2\%} \\
		\textbf{TwigVLM++} & \textbf{62.2} & \textbf{66.6} & \textbf{1773} & 57.8 & \textbf{69.0} & \textbf{80.4}  & \textbf{97.1\%} \\
	\end{tabular}
	\caption{Performance comparisons on \textbf{LLaVA-NeXT-7B}. This table follows the same layout and notes as TABLE~\ref{tab:llava1.5}.}\label{tab:llavanext}
    \vspace{-8pt}
\end{table}

\vspace{-10pt}
\subsection{Evaluation Metrics}\label{sec:eval_metric}
To evaluate the performance of TwigVLM, we consider two key aspects: \emph{accuracy preservation} and \emph{inference speedup}, which are respectively measured by the two metrics as follows. \noindent\textbf{Relative Accuracy (RelAcc)} is a widely used metric in previous works \cite{fastv,visionzip}, which calculates the proportion of the accuracy of the pruned model to the base model. When multiple benchmarks are provided, their relative accuracies are separately calculated and then averaged as the final RelAcc.
\noindent\textbf{Relative Speed (RelSpd)} quantifies the relative generation speed of a pruned model compared to its base model. The generation speed is computed by dividing the number of generated tokens by the generation time—which encompasses both the prefilling and decoding stages—and then averaged across samples from a specified benchmark.

\begin{table*}[t]
	\centering
	\setlength{\tabcolsep}{2.5pt}
	\renewcommand{\arraystretch}{1.2}
	\footnotesize
	\begin{tabular}{c|ccccccccc|ccc|cc}
	  \multirow{2}{*}{\textbf{Method}} &
	  \multicolumn{9}{c|}{\textbf{Image Benchmarks}} &
	  \multicolumn{3}{c|}{\textbf{Video Benchmarks}} &
	  \multirow{2}{*}{\textbf{RelAcc}} &
	  \multirow{2}{*}{\textbf{RelSpd}} \\
	  & \textbf{GQA} & \textbf{MME} & \textbf{MMB} & \textbf{SQA}$^{\text{I}}$ & \textbf{VQA}$^{\text{T}}$ & \textbf{VQA}$^{\text{V2}}$
	  & \textbf{MMStar} & \textbf{OCRBench} & \textbf{Blink}
	  & \textbf{VideoMME} & \textbf{EgoSchema} & \textbf{MVBench} & & \\
	  \ChangeRT{1pt}
	  \rowcolor{mygray}
	  \multicolumn{15}{c}{\textit{Upper Bound, 100\% Tokens} \ $\textbf{(100\%)}$} \\
	  \textcolor{gray}{Q2.5VL-7B} & \textcolor{gray}{60.7} & \textcolor{gray}{2347} & \textcolor{gray}{82.7} & \textcolor{gray}{75.3} & \textcolor{gray}{83.2} & \textcolor{gray}{77.9}
	  & \textcolor{gray}{63.3} & \textcolor{gray}{827} & \textcolor{gray}{56.4}
	  & \textcolor{gray}{63.4} & \textcolor{gray}{58.8} & \textcolor{gray}{69.6}
	  & \textcolor{gray}{100.0\%} & \textcolor{gray}{100.0\%} \\
	  \hline
	  \rowcolor{mygray}
	  \multicolumn{15}{c}{\textit{Retain Averaged 33.3\% Tokens} \ $\fg{(\downarrow 66.7\%)}$} \\
	  FastV & 57.2 & 2299 & 80.9 & 75.5 & 81.5 & 74.2
	  & 58.9 & 682 & 53.2
	  & 61.7 & 57.1 & 66.2
	  & 95.2\% & 101.7\% \\
	  VisionZip & 60.0 & \textbf{2387} & \textbf{83.4} & \underline{77.1} & 76.3 & 77.9
	  & \underline{61.1} & 706 & \underline{53.9}
	  & 61.5 & \underline{58.3} & \underline{68.1}
	  & 97.2\% & 103.4\% \\
	  \textbf{TwigVLM} & \textbf{60.4} & 2338 & 79.4 & \textbf{77.9} & \underline{82.6} & \underline{78.4}
	  & 60.0 & \underline{744} & 53.3
	  & \underline{62.4} & 57.0 & 68.0
	  & \underline{97.5\%} & \underline{147.7\%} \\
	  \textbf{TwigVLM++} & \underline{60.1} & \underline{2373} & \underline{82.7} & 77.0 & \textbf{83.2} & \textbf{78.9}
	  & \textbf{62.2} & \textbf{825} & \textbf{54.7}
	  & \textbf{63.6} & \textbf{58.3} & \textbf{69.1}
	  & \textbf{99.7\%} & \textbf{187.1\%} \\
	  \hline
	  \rowcolor{mygray}
	  \multicolumn{15}{c}{\textit{Retain Averaged 22.2\% Tokens} \ $\fg{(\downarrow 77.8\%)}$} \\
	  FastV & 53.5 & 2246 & 78.6 & 75.3 & 79.2 & 70.6
	  & 55.1 & 589 & 51.8
	  & 60.0 & 56.2 & 61.3
	  & 91.1\% & 103.1\% \\
	  VisionZip & \underline{59.2} & \underline{2318} & \textbf{82.6} & 77.0 & 71.6 & 76.2
	  & \underline{58.9} & 628 & \underline{53.2}
	  & 61.2 & \textbf{58.2} & \underline{66.8}
	  & 94.7\% & 105.8\% \\
	  \textbf{TwigVLM} & \textbf{59.9} & 2238 & 77.4 & \textbf{78.1} & \underline{81.4} & \underline{78.4}
	  & 57.2 & \underline{685} & 51.5
	  & \underline{61.4} & 55.1 & 66.4
	  & \underline{95.0\%} & \underline{151.2\%} \\
	  \textbf{TwigVLM++} & 59.8 & \textbf{2346} & \underline{82.1} & \underline{77.1} & \textbf{82.5} & \textbf{78.4}
	  & \textbf{60.5} & \textbf{813} & \textbf{54.0}
	  & \textbf{62.4} & \underline{58.0} & \textbf{68.6}
	  & \textbf{98.7\%} & \textbf{192.4\%} \\
	  \hline
	  \rowcolor{mygray}
	  \multicolumn{15}{c}{\textit{Retain Averaged 11.1\% Tokens} \ $\fg{(\downarrow 88.9\%)}$} \\
	  FastV & 45.1 & 1859 & 61.5 & 72.8 & 62.9 & 58.2
	  & 41.9 & 309 & 44.3
	  & 55.3 & 52.4 & 59.3
	  & 76.7\% & 104.3\% \\
	  VisionZip & 56.6 & \underline{2153} & \textbf{79.2} & \underline{76.1} & 58.7 & 70.5
	  & \underline{54.9} & 486 & \textbf{49.6}
	  & \underline{59.5} & \textbf{57.2} & \textbf{65.6}
	  & \underline{88.4\%} & 107.1\% \\
	  \textbf{TwigVLM} & \underline{57.6} & 2020 & 67.0 & 74.9 & \underline{73.0} & \textbf{75.6}
	  & 48.2 & \underline{518} & 48.6
	  & 56.6 & 49.3 & 61.0
	  & 86.0\% & \underline{152.3\%} \\
	  \textbf{TwigVLM++} & \textbf{58.3} & \textbf{2240} & \underline{78.7} & \textbf{76.1} & \textbf{79.5} & \underline{75.2}
	  & \textbf{56.6} & \textbf{772} & \underline{49.5}
	  & \textbf{60.1} & \underline{55.0} & \underline{65.5}
	  & \textbf{94.4\%} & \textbf{193.2\%} \\
	\end{tabular}
	\caption{Performance comparisons on \textbf{Qwen2.5-VL-7B} across image and video benchmarks. The vertical line separates image benchmarks (left) from video benchmarks (right). The RelAcc is evaluated following the same protocol as in TABLE~\ref{tab:llava1.5}, and the RelSpd is evaluated on MM-Vet as in Fig.~\ref{fig:main_speed}.}
	\label{tab:qwen2vl_results}
    \vspace{-8pt}
\end{table*}

\begin{figure}
    \centering
    \sbox0{\includegraphics[width=0.82\linewidth]{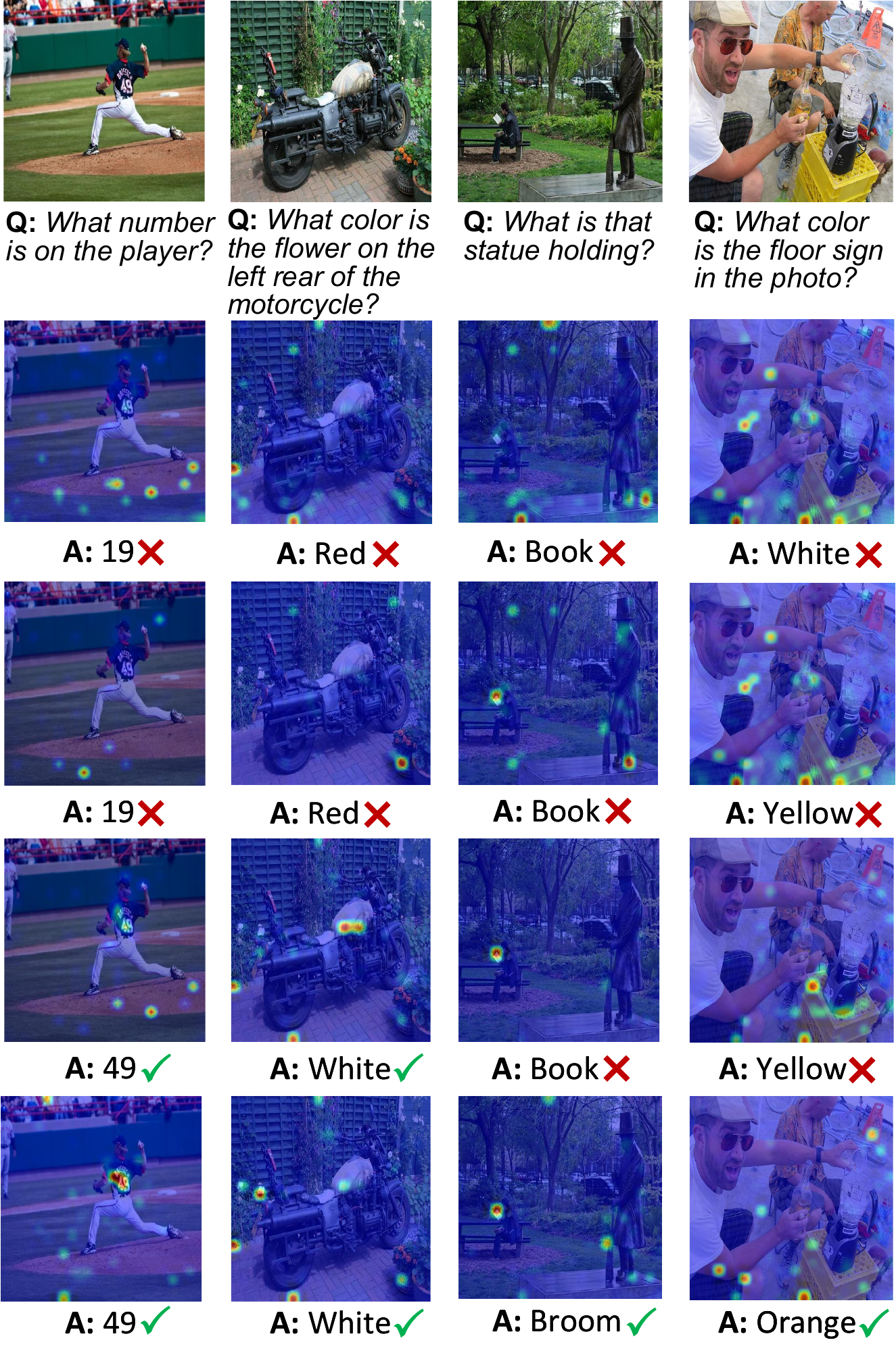}}%
    \hbox to \linewidth{%
        \vbox to \ht0{%
            \vspace{\stretch{0.35}}%
            \hbox to 0.18\linewidth{\hfill\parbox[t]{2.2cm}{\centering\scriptsize image,\\[2pt]prompt}\hfill}%
            \vspace{\stretch{1.20}}%
            \hbox to 0.18\linewidth{\hfill\scriptsize FastV \cite{fastv}\hfill}%
            \vspace{\stretch{1.0}}%
            \hbox to 0.18\linewidth{\hfill\scriptsize VisionZip \cite{visionzip}\hfill}%
            \vspace{\stretch{0.95}}%
            \hbox to 0.18\linewidth{\hfill\scriptsize \textbf{TwigVLM}\hfill}%
            \vspace{\stretch{0.95}}%
            \hbox to 0.18\linewidth{\hfill\scriptsize \textbf{TwigVLM++}\hfill}%
            \vspace{\stretch{0.6}}%
        }%
        \usebox0%
    }%
    \caption{\textbf{Visualized attention map comparisons} of TwigVLM, TwigVLM++ and two typical token pruning methods. The visualized attention maps show that our methods identify accurate visual tokens to the prompt and predict the right answer, while both counterparts fail to do that. More examples are provided in the supplementary.}\label{fig:visualization}
    \vspace{-8pt}
\end{figure}

\subsection{Main Results}
\noindent\textbf{Results on LLaVA-1.5 and LLaVA-NeXT.}
As shown in TABLE \ref{tab:llava1.5} and \ref{tab:llavanext}, we take LLaVA-1.5-7B and LLaVA-NeXT-7B as the base VLMs and compare TwigVLM with the existing VLM acceleration methods on six commonly-used VLM benchmarks.
We can see that TwigVLM consistently and significantly outperforms all its counterparts under different pruning ratios. Notably, TwigVLM achieves near-perfect performance preservation (more than 99.0\%) when pruning 66.7\% visual tokens on LLaVA-1.5-7B and 77.8\% tokens on LLaVA-NeXT-7B, showing the superiority of our TTP strategy over existing pruning strategies.
Furthermore, TwigVLM++, equipped with the RL-optimized pruning, maintains the already high accuracy at moderate pruning ratios (e.g., 99.6\% on LLaVA-1.5-7B with 66.7\% pruning), while yielding more substantial improvements under aggressive pruning---boosting RelAcc from 96.0\% to 97.7\% on LLaVA-1.5-7B and from 96.2\% to 97.1\% on LLaVA-NeXT-7B at the highest pruning ratio (88.9\%).
% Experiments on more benchmarks and VLMs (e.g., Qwen2.5-VL) are provided in the supplementary.

To better understand the effectiveness of our approach, we compare TwigVLM and TwigVLM++ with two representative methods \cite{fastv,visionzip} by visualizing the attention map used for token selection. The examples in Fig. \ref{fig:visualization} suggest that TwigVLM/TwigVLM++ can better understand the \emph{fine-grained} semantics in both the prompt and image, thus identifying more informative visual tokens for token pruning. Compared to TwigVLM, TwigVLM++ produces more accurate focal points while maintaining the necessary dispersed attention pattern.

\noindent\textbf{Results on Qwen2.5-VL for image and video understanding.}
To evaluate the generalization of TwigVLM and TwigVLM++ on stronger base VLMs, we further conduct experiments on Qwen2.5-VL-7B across both image and video benchmarks. As shown in TABLE~\ref{tab:qwen2vl_results}, TwigVLM consistently surpasses FastV and achieves comparable RelAcc with VisionZip across all pruning ratios, while TwigVLM++ substantially outperforms VisionZip by a large margin (e.g., 94.4\% \emph{vs.}\ 88.4\% at 88.9\% pruning). On video benchmarks, TwigVLM performs comparably to VisionZip, and TwigVLM++ further surpasses it in most cases, demonstrating the generalization of our approach to the video domain. Notably, the accuracy gains brought by TwigVLM++ are more pronounced on Qwen2.5-VL than on LLaVA-1.5 (e.g., +8.4\% \emph{vs.}\ +1.7\% in RelAcc at 88.9\% pruning), suggesting that the proposed multi-head twig architecture and two-stage training paradigm yield greater benefits on stronger base VLMs.

\begin{figure}
	\centering
	\includegraphics[width=1.0\linewidth]{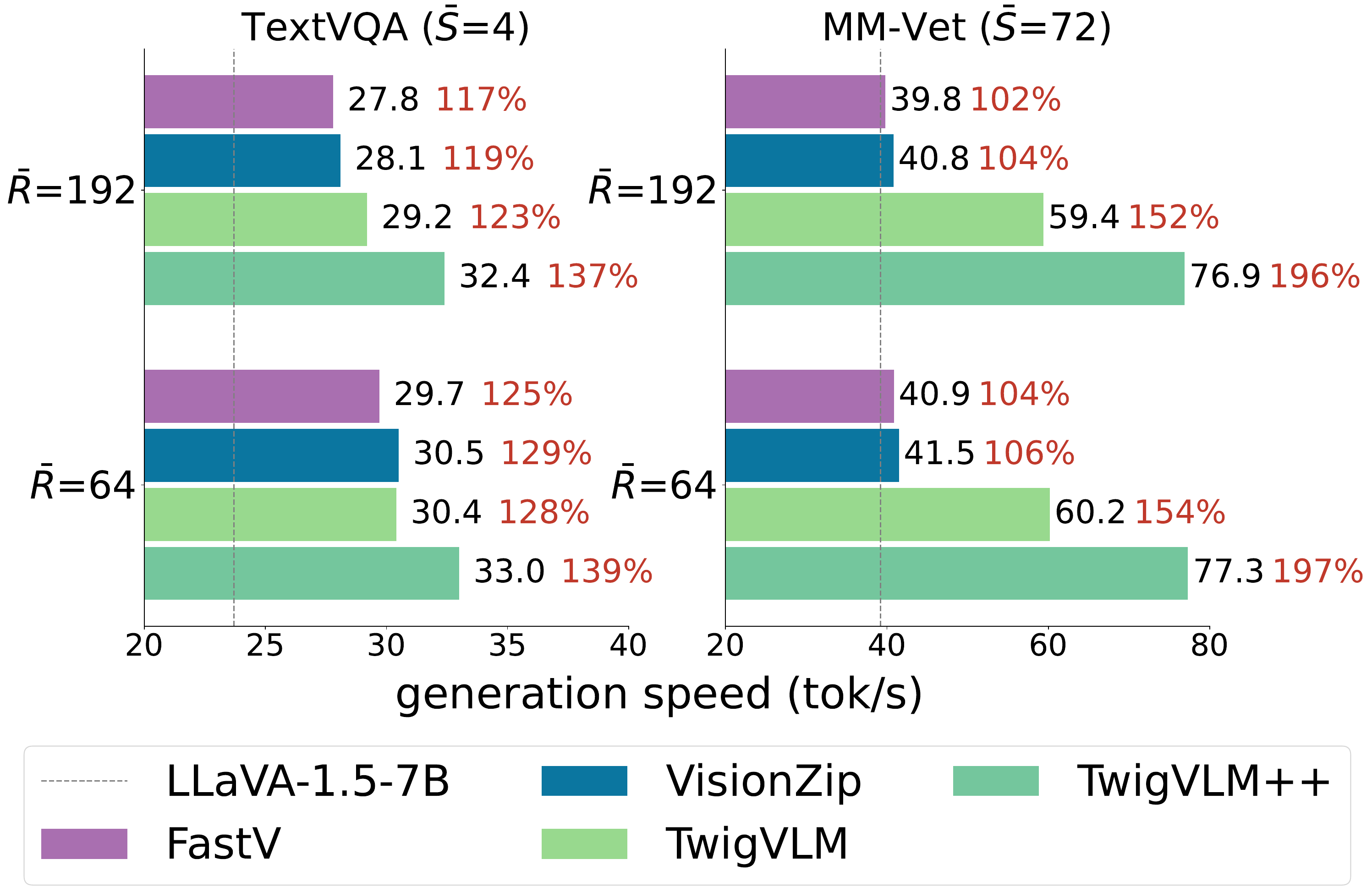}
	\caption{\textbf{Generation speed comparisons} on two benchmarks: (a) TextVQA with short responses and (b) MM-Vet with long responses. $\bar{S}$ denotes the average number of generated tokens on the whole benchmark. The \emph{RelSpd} of each bar is highlighted in \textcolor[RGB]{162,32,65}{red}.}\label{fig:main_speed}
    \vspace{-10pt}
\end{figure}

\noindent\textbf{Generation speed comparisons.} As mentioned above, TwigVLM can effectively accelerate the generation. To validate this, we conduct intensive experiments on two typical benchmarks TextVQA \cite{textvqa} and MM-Vet \cite{yu2023mmvet} to compare the speedup of different VLM acceleration methods based on LLaVA-1.5-7B. The results in Fig. \ref{fig:main_speed} show that: (i) TwigVLM achieves superior or competitive speedup in all configurations, suggesting dual advantages in speed and accuracy. (ii) All the methods attain a similar level of RelSpd (120\%$\sim$130\%) on TextVQA with short responses. In contrast, TwigVLM delivers significantly higher speedup ($\sim$150\%) than FastV ($\sim$104\%) and VisionZip ($\sim$106\%) on MM-Vet with long responses. This reveals the superiority of our SSD strategy in long response generation. (iii) A higher pruning ratio (\emph{i.e.}, a smaller $\bar{R}$) leads to more speedup on TextVQA, but has little effect on MM-Vet, which has been observed and explained in $\S$\ref{sec:pilot_prefilling}. (iv) TwigVLM++ further boosts speed across all configurations by replacing sequence-based SSD with tree-based SSD. The improvement is particularly pronounced on MM-Vet with long responses, where TwigVLM++ reaches $\sim$197\% RelSpd compared to TwigVLM's $\sim$154\%, an over 40\% additional speedup. On TextVQA with short responses, TwigVLM++ also achieves consistent gains (137\%$\sim$139\% vs.\ 123\%$\sim$128\%), confirming the effectiveness of tree-based SSD for both short and long response scenarios. Besides, the last column of TABLE~\ref{tab:qwen2vl_results} reports the RelSpd of different methods on Qwen2.5-VL-7B, where TwigVLM and TwigVLM++ consistently achieve substantially higher speedup than all counterparts, demonstrating the generalization of our acceleration strategies across different base VLMs.

\subsection{Ablation Studies for TwigVLM}
To validate the effectiveness of TwigVLM's key components, we conduct ablation experiments using the default setting ($T$=3, $K$=2, $K_\text{f}$=24, $\bar{R}$=64). Taking LLaVA-1.5-7B as the reference model, the relative accuracy (RelAcc) is evaluated on the six benchmarks mentioned in TABLE \ref{tab:llava1.5} and the relative speed (RelSpd) is evaluated on MM-Vet. Results in TABLE \ref{tab:abla} are discussed in detail below.

\vspace{3pt}
\noindent\textbf{Visual token selection.} The performance of visual token pruning relies on the chosen attention map for token selection. TABLE \ref{tab:selection_source} compares three TwigVLM variants with token pruning guided by different attention sources: (a) the $K$-th layer of the VLM backbone, (b) the ($K$+$T$)-th layer of the VLM backbone, and (c) the last twig layer which has the same depth as (b). From the results, we can see that using the attention map of the $K$-th backbone layer yields the lowest accuracy (82.3\%), which verifies the conclusion in $\S$\ref{sec:pilot_prefilling} that the semantics in early-layer attention is ambiguous. With the same attention depth $D$=($K$+$T$), selecting tokens by the attention from the backbone (86.2\%) is inferior to that from the twig layer (96.0\%), indicating that the quality of the attention map is also impacted by the ``distance'' between the layer it is on and the prediction head.

\begin{table*}
	\renewcommand\arraystretch{1.2}
	\scriptsize
	\begin{subtable}[t]{0.3\textwidth}
		\centering
		\begin{tabular}{lc}
			attention source, depth $D$~~~~~ & \makecell{RelAcc (\%)}\\
			\ChangeRT{1.0pt}
			(a) VLM backbone, $K$               & 82.3\\
			(b) VLM backbone, $K$+$T$               & 86.2\\
			\multicolumn{1}{>{\columncolor{twiggreen}}l}{(c) twig layer, $K$+$T$}                     & \textbf{96.0}\\
		\end{tabular}
		\captionsetup{font=footnotesize}
		\subcaption{\textbf{Visual token selection.} The attention map from the last twig layer results in the highest accuracy as it is close to the prediction head.}
		\label{tab:selection_source}
	\end{subtable}
	\quad
	\renewcommand\arraystretch{1.2}
	\scriptsize
	\begin{subtable}[t]{0.3\textwidth}
		\centering
		\begin{tabular}{lc}
			acceleration strategy & \makecell{RelSpd (\%)}\\
			\ChangeRT{1.0pt}
			(a) token purning (FastV \cite{fastv})          & 104.3\\
			(b) SSD      & 146.7\\
			\multicolumn{1}{>{\columncolor{twiggreen}}l}{(c) TTP \& SSD}    & \textbf{153.6}\\
		\end{tabular}
	\captionsetup{font=footnotesize}
	\subcaption{\textbf{Acceleration strategies.} The TTP and SSD strategies are complementary to each other, and their synergy achieves the highest speedup.}
	\label{tab:accelaration}
\end{subtable}
\quad
\renewcommand\arraystretch{1.2}
\scriptsize
\begin{subtable}[t]{0.33\textwidth}
	\centering
	\begin{tabular}{lcc}
		initialization strategy & RelAcc (\%) & RelSpd (\%)\\
		\ChangeRT{1.0pt}
		(a) random init. & 87.2 & 120.4\\
		(b) VLM layers[$L$-$T$:$L$] & 90.4 & 131.4\\
		\multicolumn{1}{>{\columncolor{twiggreen}}l}{(c) VLM layers[$K$:$K$+$T$]} & \textbf{96.0} & \textbf{153.6}\\
	\end{tabular}
	\captionsetup{font=footnotesize}
	\subcaption{\textbf{Twig block initialization.} The twig layers initialized from the $K$-th to ($K$+$T$)-th layers in the base VLM achieve the best accuracy and speed.}
	\label{tab:initialization}
\end{subtable}
\\[1.em]
\renewcommand\arraystretch{1.2}
\scriptsize
\begin{subtable}[t]{0.3\textwidth}
	\centering
	\begin{tabular}{ccc}
		~~~~~$T$~~~~~ & ~~RelAcc (\%)~~& ~~RelSpd (\%)~~\\
		\ChangeRT{1.0pt}
		1  & 93.9& \textbf{154.1}\\
		2  & 95.2& 152.6\\
		\multicolumn{1}{>{\columncolor{twiggreen}}c}{3} & \textbf{96.0} & 153.6\\
		4  & 95.8 & 145.4\\
	\end{tabular}
	\captionsetup{font=footnotesize}
	\subcaption{\textbf{Number of twig layers.} $T$=3 results in the highest accuracy with negligible speed degradation compared to $T$=1.}
	\label{tab:twig_layers}
\end{subtable}
\quad
\renewcommand\arraystretch{1.2}
\scriptsize
\begin{subtable}[t]{0.3\textwidth}
	\centering
	\begin{tabular}{cccc}
		~~$K$~~ & ~~$R$~~ & RelAcc (\%) & RelSpd (\%)\\
		\ChangeRT{1.0pt}
		0& 85  & 93.6 & 137.5\\
		1& 64  & 95.5 & 136.5\\
		\multicolumn{1}{>{\columncolor{twiggreen}}c}{2} &  \multicolumn{1}{>{\columncolor{twiggreen}}c}{41}  & \textbf{96.0} & \textbf{153.6}\\
		3& 15  & 85.5 & 145.2\\
	\end{tabular}
	\captionsetup{font=footnotesize}
	\subcaption{\textbf{Pruning position.}  When $\bar{R}$ is fixed to 64, $K$=2 achieves the optimal balance between accuracy and speed.}
	\label{tab:pruning_position}
\end{subtable}
\quad
\renewcommand\arraystretch{1.2}
\scriptsize
\begin{subtable}[t]{0.35\textwidth}
	\centering
	\begin{tabular}{ccccc}
		FinalWipe & $K_\text{f}$ & $R$ & RelAcc (\%) & RelSpd (\%)\\
		\ChangeRT{1.0pt}
		$\times$   & 32 & 30  & 93.1 & \textbf{154.6}\\
		\checkmark & 20 & 50  & 95.8 & 151.3\\
		\multicolumn{1}{>{\columncolor{twiggreen}}c}{\checkmark} &\multicolumn{1}{>{\columncolor{twiggreen}}c}{24} &  \multicolumn{1}{>{\columncolor{twiggreen}}c}{41}  & \textbf{96.0} & 153.6\\
		\checkmark & 28 & 37  & 95.1 & {154.1}\\
	\end{tabular}
	\captionsetup{font=footnotesize}
	\subcaption{\textbf{FinalWipe.} The FinalWipe strategy facilitates model accuracy and $K_\text{f}$=24 is the optimal choice to balance accuracy and speed.}
	\label{tab:finaldrop}
\end{subtable}
\caption{\textbf{Ablation experiments for TwigVLM}. Taking LLaVA-1.5-7B as the base VLM, the relative accuracy (RelAcc) is evaluated on the six benchmarks mentioned in TABLE \ref{tab:llava1.5} and the relative speed (RelSpd) is evaluated on MM-Vet. In each table, the best result is \textbf{bolded} and the default setting ($T$=3, $K$=2, $K_\text{f}$=24, $\bar{R}$=64) is marked in \colorbox{twiggreen}{green}. }
\label{tab:abla}
\end{table*}

\vspace{3pt}
\noindent\textbf{Acceleration strategies.} The inference acceleration of TwigVLM is achieved by the synergy between the twig-guided token pruning (TTP) and self-speculative decoding (SSD) strategies during prefilling and decoding, respectively. In TABLE \ref{tab:accelaration}, we compare the three acceleration strategies as follows: (a) visual token pruning by FastV \cite{fastv}, (b) SSD only without token pruning, and (c) the standard TwigVLM with SSD and TTP. While FastV achieves modest speedup (104.3\%), SDD delivers a remarkable improvement (146.7\%) as the decoding time is longer than prefilling. Finally, the TTP and SSD strategies are complementary and their synergy achieves the highest speedup (153.6\%).

\vspace{3pt}
\noindent\textbf{Twig block initialization.} The initialization strategy for twig layers has a prominent impact on TwigVLM's performance. From TABLE \ref{tab:initialization}, we can see that: random initialization in (a) performs poorly due to the lack of knowledge inherited from the base VLM. Moreover, the strategy in (b) that uses the last $T$ VLM layers to initialize the twig layers results in distinct improvements. Finally, initializing from layers $K$ to $K+T$ achieves the best results in both accuracy and speed.
The speed improvements are contributed by the higher token acceptance rate\footnote{The token acceptance rate measures how often verification accepts each of the draft tokens \cite{layerskip}.}. More detailed analyses about token acceptance rate are provided in the supplementary.

\vspace{3pt}
\noindent\textbf{Number of twig layers.} TABLE \ref{tab:twig_layers} varies the number of twig layers $T$ from 1 to 4 to investigate its effect on TwigVLM. We observe that a single twig layer ($T$=1) leads to the highest speed but the lowest accuracy. Increasing $T$ to 2 or 3 improves accuracy with negligible speed degradation. However, further increasing $T$ to 4 leads to a distinct performance drop in terms of both accuracy and speed.
These observations can be explained by two factors: i) improving retained token selection (via more twig layers) obtains diminishing accuracy returns, which finally saturates at $T$=3, and ii) deeper twig block involves linearly increasing computation while the token acceptance ratio in SSD achieves saturation, leading to a decrease in generation speed. Thus, $T$=3 offers an optimal choice for the number of twig layers.

\vspace{3pt}
\noindent\textbf{Pruning position.} To achieve an optimal number of retained tokens when $\bar{R}$ is fixed to 64, we use Eq.(\ref{eq:avgR_finalwipe}) to explore different combinations of $K$ and $R$. Increasing $K$ leads to a smaller $R$, which means a trade-off between the number of visual tokens in early layers and the number of visual tokens in subsequent layers. As shown in TABLE \ref{tab:pruning_position}, both the accuracy and speed achieve the best results at $K$=2, which is chosen in our default setting.

\vspace{3pt}
\noindent\textbf{FinalWipe.} The FinalWipe works by removing all visual tokens after the $K_\text{f}$-th layer. TABLE \ref{tab:finaldrop} ablates different TwigVLM variants with/without this strategy. When FinalWipe is not introduced, the accuracy is relatively low due to the insufficiency of retained tokens. Introducing FinalWipe facilitates accuracy with limited speed drop, which can be explained by the fact that the visual tokens become less important in late layers. $K_\text{f}$=24 achieves a good balance between accuracy and speed, thus is used as the default.

\begin{table}
    \centering
    \renewcommand\arraystretch{1.2}
	\footnotesize
    \setlength{\tabcolsep}{4.5pt}
    \begin{subtable}[t]{0.95\columnwidth}
        \centering
        \begin{tabular}{l l c c}
        	% \toprule
        	Head(s) & Loss & Stage-1 & Stage-2 \\
        	\ChangeRT{1.0pt}
        	% \midrule
        	D-Head     & $\mathcal{L}_\text{NTP}$                    & 96.0\% & - \\
        	D-Head + P-Head & $\mathcal{L}_\text{NTP}+\mathcal{L}_\text{AttnKL}$           & 95.0\% & 97.2\% \\
        	\multicolumn{1}{>{\columncolor{twiggreenplus}}l}{D-Head + P-Head} & \multicolumn{1}{>{\columncolor{twiggreenplus}}l}{$\mathcal{L}_\text{NTP}+\mathcal{L}_\text{AttnKL}+\mathcal{L}_\text{PredKL}$}  & \textbf{96.4}\% & \textbf{97.7}\% \\
        	% \bottomrule
        \end{tabular}
        \caption{Different training settings in stage-1. Scores refer to the RelAcc evaluated with $\bar{R}$=64.}
    \end{subtable}
    \vspace{5pt}
    
    \setlength{\tabcolsep}{4.5pt}
	\renewcommand\arraystretch{1.2}
	\footnotesize
    \begin{subtable}[t]{0.48\columnwidth}
        \centering
        \begin{tabular}{c c c c}
            % \multirow{2}{*}{$\bar{R}$\,@\,RL} & \multicolumn{3}{c}{RelAcc(\%) @ $\bar{R}$} \\
            \multirow{2}{*}{\makecell{RL setting\\for $\bar{R}$}} & \multicolumn{3}{c}{RelAcc(\%) @ $\bar{R}$} \\
            \cline{2-4}
            & 192 & 128 & 64 \\
            \ChangeRT{1.0pt}
            static $\bar{R}$=192     & 99.4 & \textbf{99.3} & 97.1 \\
            static $\bar{R}$=128     & 99.5 & 99.2 & 97.4 \\
            static $\bar{R}$=64      & 99.1 & 99.0 & \textbf{98.0} \\
            \multicolumn{1}{>{\columncolor{twiggreenplus}}c}{dynamic}                  & \textbf{99.6} & 99.2 & 97.7 \\
        \end{tabular}
		\caption{Different pruning-ratio strategy in stage-2.}
        \label{tab:left_part}
    \end{subtable}
    %\hfill
    \setlength{\tabcolsep}{4.5pt}
	\renewcommand\arraystretch{1.2}
	\footnotesize
    \begin{subtable}[t]{0.48\columnwidth}
        \centering
        \begin{tabular}{c c c c}
            \multirow{2}{*}{\makecell{\#\,training \\ samples}} & \multicolumn{3}{c}{RelAcc(\%) @ $\bar{R}$} \\
            \cline{2-4}
            & 192 & 128 & 64 \\
            \ChangeRT{1.0pt}
            10k  & 99.1 & 98.8 & 97.2 \\
            20k  & 99.4 & 99.1 & 97.6 \\
            \multicolumn{1}{>{\columncolor{twiggreenplus}}c}{50k} & \textbf{99.6} & \textbf{99.2} & 97.7 \\
            100k & 99.4 & 99.0 & \textbf{97.8} \\
        \end{tabular}
		\caption{Different sizes of the training data in stage-2.}
        \label{tab:right_part}
    \end{subtable}
	\caption{\textbf{Ablation experiments for TwigVLM++}. The best result is \textbf{bolded} and the default setting is marked in \colorbox{twiggreenplus}{dark green}.}
    \label{tab:stage2_ablation}
    \vspace{-10pt}
\end{table}

\subsection{Ablation Studies for TwigVLM++}

We ablate the key design choices in TwigVLM++ training on LLaVA-1.5-7B. Results are reported in TABLE \ref{tab:stage2_ablation}.

\vspace{3pt}
\noindent\textbf{Architecture and loss in stage-1.}
TABLE \ref{tab:stage2_ablation}a compares different head configurations and loss combinations. With the single D-Head baseline achieving 96.0\% RelAcc after stage-1, introducing the P-Head with only $\mathcal{L}_\text{AttnKL}$ actually degrades stage-1 performance to 95.0\%, suggesting that the multi-head architecture splits the training capacity and leads to insufficient optimization of each head. Adding $\mathcal{L}_\text{PredKL}$ effectively compensates for this issue, raising stage-1 accuracy to 96.4\% and yielding the best stage-2 result of 97.7\%. This confirms that strong-to-weak distillation provides complementary supervision that benefits both the D-Head and P-Head training.

\vspace{3pt}
\noindent\textbf{Pruning-ratio strategy in stage-2.}
TABLE \ref{tab:stage2_ablation}b compares static and dynamic pruning-ratio strategies for stage-2 RL. Training with a static $\bar{R}$=64 achieves the highest accuracy at $\bar{R}$=64 (98.0\%) but degrades performance at larger retention settings ($\bar{R}$=192: 99.1\%, $\bar{R}$=128: 99.0\%), as the policy overfits to aggressive pruning. The reverse trend holds for static $\bar{R}$=192. In contrast, the dynamic strategy achieves the best result at $\bar{R}$=192 (99.6\%) while maintaining competitive accuracy across other ratios, delivering the most balanced performance overall.

\vspace{3pt}
\noindent\textbf{Training data scale in stage-2.}
TABLE \ref{tab:stage2_ablation}c varies the number of RL training samples drawn from the LLaVA-665K SFT dataset. Performance improves notably from 10k to 20k samples, after which the gains largely saturate---the gap between 20k and 50k is marginal (e.g., 97.6\% vs.\ 97.7\% at $\bar{R}$=64), and scaling to 100k yields no further improvement. This indicates that only 50k samples suffice for the RL training to converge, demonstrating the data efficiency of the proposed stage-2 optimization.
\vspace{-8pt}

\section{Conclusion}
In this paper, we present TwigVLM---a conceptually simple yet effective approach that accelerates VLM inference by growing a lightweight twig block upon an early layer of the base VLM, enabling both twig-guided token pruning (TTP) for prefilling acceleration and self-speculative decoding (SSD) for decoding acceleration. To further improve the pruning quality, we extend TwigVLM to TwigVLM++ by introducing a multi-head twig architecture that decouples pruning from next-token prediction, and a two-stage training paradigm that combines distillation learning with reinforcement learning to directly optimize the pruning decisions. Furthermore, the tree-based SSD is adopted to achieve higher decoding throughput. Extensive ablations, comparative experiments, and comprehensive analyses on a wide range of image and video benchmarks demonstrate the superiority of TwigVLM and TwigVLM++ over existing state-of-the-art methods. We hope our work may serve as a solid baseline to inspire future research on visual token reduction and VLM acceleration.

\appendices

\section{More Implementation Details}\label{sec:appendix_implementation}
\noindent\textbf{TwigVLM/TwigVLM++ training.} As described in the main text, the twig block is trained by finetuning the shallow VLM $\mathcal{M}_s$. Specifically, $\mathcal{M}_s$ is initialized with the weights of the first $K$+$T$ layers and the prediction head of the corresponding base VLM $\mathcal{M}_b$. During finetuning, only the last $T$ layers and the prediction head—collectively termed the twig block—are updated, while the first $K$ layers remain frozen. This process follows the same training manner to train the base VLM $\mathcal{M}_b$. Theoretically, any suitable multimodal instruction tuning dataset can be employed to finetune $\mathcal{M}_s$.

\vspace{2pt}
The base VLMs evaluated in the main text experiments all leverage the open-source datasets to train their twig blocks.  Specifically, we use the LLaVA-665K dataset \cite{llava1.5} to train TwigVLM/TwigVLM++ for LLaVA-1.5-7B and LLaVA-NeXT-7B models, and a dataset of 5M single-image samples from the MAmmoTH-VL-10M dataset \cite{mammoth-VL} for Qwen2.5-VL-7B. The optimization hyper-parameters used for training TwigVLM/TwigVLM++ are detailed in TABLE~\ref{tab:hyperparameters}. All training is performed on a server equipped with 8 NVIDIA A100 GPUs. Under these conditions, the training of TwigVLM is highly efficient, requiring only approximately 10\% of the time needed to train the corresponding base VLM, \emph{e.g.}, training the twig block for the LLaVA-1.5-7B model takes about 10 GPU hours, while the training of the original LLaVA-1.5-7B takes about 100 GPU hours. TwigVLM++ training also needs only 20\% of that time.

\begin{table}[h]
	\centering
    \renewcommand\arraystretch{1.2}
	\small
	\begin{tabular}{l|cc}
		config & setting \\
		\ChangeRT{1pt}
		\rowcolor{mygray}
		\multicolumn{2}{c}{\textit{TwigVLM \& TwigVLM++ stage-1 (shared)}} \\
		optimizer & AdamW \\
		weight decay & 0. \\
		optimizer momentum & $\beta_1, \beta_2{=}0.9, 0.98$ \\
		batch size & 128 \\
		learning rate schedule & cosine decay \\
		peak learning rate & 1e-4 \\
		warm-up strategy & linearly warm-up\\
		warm-up ratio & 0.03 \\
		training samples & 665K \\
		training epochs & 1 \\
        \hline
		\rowcolor{mygray}
		\multicolumn{2}{c}{\textit{TwigVLM++ stage-1 (additional)}} \\
		PredKL coefficient $\alpha$ & 0.1 \\
		AttnKL coefficient $\gamma$ & 1.0 \\
		\hline
		\rowcolor{mygray}
		\multicolumn{2}{c}{\textit{TwigVLM++ stage-2}} \\
		optimizer & AdamW \\
		peak learning rate & 2e-5 \\
		batch size & 128 \\
		group size $G$ & 32 \\
		candidate set $\mathcal{R}$ & \{64,85,107,128,149,171,192\} \\
		annealing params $(\beta_{\max}, p)$ & (8.0, 2.0) \\
		training samples & 50K \\
		training epochs & 1 \\
	\end{tabular}
	\caption{\textbf{Training settings.} The first section lists hyper-parameters shared by TwigVLM and the training stage-1 of TwigVLM++. The second and third sections list additional hyper-parameters specific to TwigVLM++.} \label{tab:hyperparameters}
\end{table}

\vspace{3pt}
\noindent\textbf{Twig-guided token pruning (TTP).} During inference, TwigVLM leverages the TTP strategy to perform token pruning over the base VLM: (i) at the $K$-th layer, selecting $R$ key visual tokens (output by the $K$-th layer) and discarding the rest tokens guided by the attention map from the last twig layer, and (ii) applying the FinalWipe strategy to further remove all the visual tokens after the $K_\text{f}$-th layer. Therefore, we adjust the value of $R$ to satisfy different pruning ratios calculated by the average number of retained visual tokens $\bar{R}$. TABLE~\ref{tab:pruning_setting} shows the default pruning settings for TwigVLM under different pruning ratios.

\begin{table}[t]
	%\vspace{-.5em}
	\centering
    \renewcommand\arraystretch{1.2}
	\small
	\begin{tabular}{cc|ccc}
		$\bar{R}$ & pruning ratio &$K$ & $R$ & $K_\text{f}$\\
		\ChangeRT{1pt}
		192 & 66.7\% & 2 & 227 & 24  \\
		128 & 77.8\% & 2 & 134 & 24 \\
		64 & 88.9\% & 2 & 41 & 24  \\
	\end{tabular}
	\vspace{-5pt}
	\caption{\textbf{Pruning settings.} These hyper-parameters correspond to the default TTP settings of different pruning ratios.} \label{tab:pruning_setting}
	%\vspace{-10pt}
\end{table}

\vspace{3pt}
\noindent\textbf{Self-speculative decoding (SSD).} For efficient generation of long responses, TwigVLM applies the SSD strategy by using the $\mathcal{M}_s$ as the \emph{draft} model and $\mathcal{M}_b$ as the \emph{target} model. Specifically, in each SSD iteration, the draft model efficiently predicts $\delta=5$ subsequent draft tokens in an autoregressive manner. To further improve efficiency, this draft generation process is equipped with an early-exit mechanism that allows the draft model to stop generation if the probability of the current predicted token falls below a predefined threshold $\theta=0.6$. The target model then verifies these generated draft tokens in parallel, accepts those matching the target model's predictions, and then predicts a next token by itself. The iteration repeats until the \texttt{<EOS>} token is generated. Note that the TTP and SSD strategies can be seamlessly integrated, as detailed in Algorithm~\ref{alg:inference}.

\noindent\textbf{Tree-based self-speculative decoding.} TwigVLM++ replaces the sequential SSD with a tree-based variant to increase the number of accepted tokens per verification step. The token tree $\mathcal{T}$ is rooted at the last accepted token and constructed level by level, governed by three hyperparameters: an expansion width $\mathcal{E}$, a selection width $\mathcal{K}$, and a tree depth $\mathcal{D}$. Let $\mathcal{T}_l$ denote the set of nodes at level $l$. At the first level, $\mathcal{M}_s$ computes the draft distribution conditioned on the current prefix and takes the top-$\mathcal{E}$ tokens as children of the root, giving $|\mathcal{T}_1|{=}\mathcal{E}$. For each subsequent level $l>1$, the top-$\mathcal{K}$ nodes from $\mathcal{T}_{l-1}$ (ranked by prediction probability) are selected for expansion; each selected node $u$ is fed into $\mathcal{M}_s$ together with its root-to-$u$ prefix to produce $\mathcal{E}$ children, yielding $|\mathcal{T}_l|{=}\mathcal{K}{\cdot}\mathcal{E}$. To bound the verification cost, the completed tree is pruned to retain at most $N_{\max}$ candidate nodes by preferentially keeping the highest-confidence leaf nodes and their ancestors. In our default setting, we use $\mathcal{E}{=}10$, $\mathcal{K}{=}10$, $\mathcal{D}{=}4$, and $N_{\max}{=}60$. For verification, $\mathcal{M}_b$ processes the pruned tree in a single forward pass via \emph{tree attention}~\cite{specinfer}, using a \emph{topology-aware causal mask} so that each node attends only to its ancestors. The target model traverses the tree from the root, accepting a child at each level whose token matches the target model's prediction, and stops at the first level where no child matches; a bonus token predicted by $\mathcal{M}_b$ is then appended. The full procedure is detailed in Algorithm~\ref{alg:inference_twigpp}.

\begin{algorithm}[t]
    \caption{Pseudocode of TwigVLM's inference process}
    \label{alg:inference}
    % \algcomment{\fontsize{7.2pt}{0em}\selectfont \texttt{concat}: concatenation operation; \texttt{argmax}: selects the index with the highest probability; \texttt{pruning}: removes visual tokens based on attention scores.}
    \definecolor{codeblue}{rgb}{0.25,0.5,0.5}
    \lstset{
      backgroundcolor=\color{white},
      basicstyle=\fontsize{7.2pt}{7.2pt}\ttfamily\selectfont,
      columns=fullflexible,
      breaklines=true,
      captionpos=b,
      commentstyle=\fontsize{7.2pt}{7.2pt}\color{codeblue},
      keywordstyle=\fontsize{7.2pt}{7.2pt},
    }
    \begin{lstlisting}[language=python]
# bVLM: the base VLM model, i.e., M_b
# twig: the twig block
# K: Number of shared low layers
# K_f: The position to apply FinalWipe
# R: Number of retained visual tokens when pruning
# delta: Maximum draft token length
# theta: Confidence threshold to stop draft

def sVLM_forward(tokens):
    X_k = bVLM.forward_low_layers(tokens, k=K)
    prob, Attn_last = twig.forward(X_k)
    a_i = argmax(prob)
    return X_k, prob, Attn_last, a_i

    
def TwigVLM_inference(img, ques):
    draft_toks = [] # temporary buffer for draft tokens
    final_resp = [] # buffer for final response
    
    # Prefilling stage of sVLM 
    X_k, _, Attn, a_i = sVLM_forward((img, ques))
    draft_toks.append(a_i)
    
    # Prune visual tokens in X_k using Eq. (5)
    # X_k_b means shared token latents for bVLM
    X_k_b = pruning(X_k, Attn, r=R)
    
    # The loop of self speculative decoding
    while EOS_TOKEN not in final_resp:
        X_k, prob, _, a_i = sVLM_forward(a_i)
        draft_toks.append(a_i)
        X_k_b = concat(X_k_b, X_k, axis=1)

        # the condition to stop draft and verify
        if len(draft_toks) >= delta or prob < theta:
            # removing all visual tokens after layer K_f
            tgt_probs = bVLM.forward_high_layers(
                X_k_b, k=K, fianl_wipe=K_f)
            # verification
            right_toks = [a for a, p in zip(draft_toks, tgt_probs[:-1]) if argmax(p) == a]
            right_toks.append(argmax(tgt_probs[-1]))
            final_resp.extend(right_tokens)
            # reset temporary variables
            draft_toks = []
            X_k_b = None
            a_i = final_resp[-1]
    return final_resp
      \end{lstlisting}
    %   # For first the forward pass (prefilling stage), 
      % # (image encoding omitted for simplicity)
      % right_tokens = [a for a, prob in \
      %               zip(draft_tokens, probs) if argmax(prob) == a]
      % # get the token prediction from the twig and attention
      % # weights (softmax(QK)) from the last twig layer
      % # Clear KV-cache and reset temporary variables
      % wrong_len = len(draft_tokens) - len(right_tokens)
      % bVLM.clear_wrong_kvcache(wrong_len)
      % twig.clear_wrong_kvcache(wrong_len)

    %   def verify(draft_tokens, target_probs):
    % right_tokens = []
    % for i in range(len(draft_tokens)):
    %     if argmax(target_probs[i]) == draft_tokens[i]:
    %         right_tokens.append(draft_tokens[i])
    %     else:
    %         break
    % # add the additional token predicted by the bVLM
    % right_tokens.append(argmax(target_probs[i]))
    % return right_tokens
    % \vspace{-10pt}
    \end{algorithm}
% \vspace{-0.5cm}
\begin{algorithm}[t]
    \caption{Pseudocode of TwigVLM++'s inference process}
    \label{alg:inference_twigpp}
    \definecolor{codeblue}{rgb}{0.25,0.5,0.5}
    \lstset{
      backgroundcolor=\color{white},
      basicstyle=\fontsize{7.2pt}{7.2pt}\ttfamily\selectfont,
      columns=fullflexible,
      breaklines=true,
      captionpos=b,
      commentstyle=\fontsize{7.2pt}{7.2pt}\color{codeblue},
      keywordstyle=\fontsize{7.2pt}{7.2pt},
    }
    \begin{lstlisting}[language=python]
# bVLM: the base VLM model, i.e., M_b
# twig: the twig block
# K: Number of shared low layers
# K_f: The position to apply FinalWipe
# R: Number of retained visual tokens when pruning
# delta: Number of tree draft iterations
# top_k: Branch factor for tree-based drafting
# max_tokens: Maximum number of candidate tokens in tree

def sVLM_forward(tokens, tree_mask):
    X_k = bVLM.forward_low_layers(tokens, k=K)
    prob, X_t, qk = twig.forward(X_k, tree_mask)
    return X_k, X_t, prob, qk
    
def TwigVLM++_inference(img, ques):
    final_resp = []  # buffer for final response

    # Prefilling stage of sVLM
    X_k, X_t, _, qk = sVLM_forward((img, ques), None)

    # Prune visual tokens using P-Head
    Attn_phead = P_Head(X_t, qk, img_tags)
    X_k_b = pruning(X_k, Attn_phead, r=R)

    # The loop of tree-based speculative decoding
    while EOS_TOKEN not in final_resp:
        # Tree-based draft generation
        draft_toks = []
        for i in range(delta):
            X_k, prob, _, _ = sVLM_forward(a_i, tree_mask)
            # Top-k sampling for tree expansion
            topk_toks = topk(prob, k=top_k)
            draft_toks.append(topk_toks)
            a_i = topk_toks
            X_k_b = concat(X_k_b, X_k, axis=1)

        # Build tree structure from draft tokens
        tree_cands, tree_pos_ids, tree_mask = \
            build_tree_candidates(draft_toks, top_k, max_tokens)

        # Verify with tree-structured forward
        tgt_probs = bVLM.forward_high_layers(
            X_k_b, tree_cands, tree_pos_ids, tree_mask,
            k=K, final_wipe=K_f)

        # Select best candidate path
        accept_len = verify_tree(tree_cands, tgt_probs)
        best_path = select_best_path(tree_cands, accept_len)

        # Accept tokens and add bonus token
        accept_toks = best_path[:accept_len]
        accept_toks.append(argmax(tgt_probs[accept_len]))
        final_resp.extend(accept_toks)

        # Reset for next iteration
        a_i = final_resp[-1]
        X_k_b = None

    return final_resp
    \end{lstlisting}
    \end{algorithm}

\begin{table}[t]
\centering
\setlength{\tabcolsep}{2.0pt}
\renewcommand{\arraystretch}{1.3}
\footnotesize
	\begin{tabular}{c | c c c c c c | c }
		\textbf{Method} & \textbf{GQA} & \textbf{MMB} & \textbf{MME} & \textbf{VQA}$^{\text{T}}$ & \textbf{SQA}$^{\text{I}}$ & \textbf{VQA}$^{\text{V2}}$ &  \makecell[c]{\textbf{RelAcc}} \\
		\ChangeRT{1pt}
		\rowcolor{mygray}
		\multicolumn{8}{c}{\textit{Upper Bound, 576 Tokens} \ $\textbf{(100\%)}$} \\
		\textcolor{gray}{LLaVA-1.5-13B} & \textcolor{gray}{63.2} & \textcolor{gray}{67.7} & \textcolor{gray}{1818} & \textcolor{gray}{61.3}  & \textcolor{gray}{72.8} & \textcolor{gray}{80.0} &  \textcolor{gray}{100\%} \\
		\hline
		\rowcolor{mygray}
		\multicolumn{8}{c}{\textit{Retain Averaged 192 Tokens} \ $\fg{(\downarrow 66.7\%)}$} \\
		FastV       & 60.3 & 67.4 & 1807 & 60.4 & \textbf{74.0} & 77.7  & 98.6\% \\
		VisionZip       & 59.1 & 66.9 & 1754 & 59.5 & 73.5 & 78.1  & 97.4\% \\
		VisionZip\ddag & 61.6 & 67.1 & 1790 & 59.9 & 72.7 & 78.6  & 98.5\% \\
		\textbf{TwigVLM} & 62.5 & \textbf{68.6} & \textbf{1840} & 60.4 & 73.1 & 79.4 & \textbf{99.9\%} \\
        \textbf{TwigVLM++} & \textbf{62.6} & 68.2 & 1829 & \textbf{60.4} & \textbf{73.2} & \textbf{79.4} & 99.8\% \\
		\hline
		\rowcolor{mygray}
		\multicolumn{8}{c}{\textit{Retain Averaged 128 Tokens} \ $\fg{(\downarrow 77.8\%)}$} \\
		FastV       & 57.5 & 65.9 & 1758 & 58 & 73.8 & 74.3  & 95.7\% \\
		VisionZip       & 57.9 & 66.7 & 1743 & 58.7 & \textbf{74.0} & 76.8  & 96.6\% \\
		VisionZip\ddag & 60.1 & \textbf{67.6} & 1736 & 59.2 & 73.0 & 77.6  & 97.4\% \\
		\textbf{TwigVLM}  & 61.2 & 66.9 & 1811 & 60.2 & 73.4 & 79.1  & 98.9\% \\
        \textbf{TwigVLM++} & \textbf{62.3} & \textbf{67.7} & \textbf{1849} & \textbf{60.3} & 72.9 & \textbf{79.1} & \textbf{99.6\%} \\
		\hline
		\rowcolor{mygray}
		\multicolumn{8}{c}{\textit{Retain Averaged 64 Tokens} \ $\fg{(\downarrow 88.9\%)}$} \\
		FastV       & 50.1 & 55.9 & 1408 & 52.2 & 73.2 & 61.1  & 83.6\% \\
		VisionZip       & 56.2 & 64.9 & 1676 & 57.4 & \textbf{74.4} & 73.7  & 94.2\% \\
		VisionZip\ddag & 58.1 & 65.6 & 1671 & 58.5 & 72.3 & 75.2  & 94.9\% \\
		\textbf{TwigVLM}  & 60.0 & \textbf{67.4} & 1765 & 58.4 & 72.4 & 77.0  & 97.1\% \\
        \textbf{TwigVLM++} & \textbf{61} & 65.7 & \textbf{1793} & \textbf{58.5} & 72.7 & \textbf{78.0} & \textbf{97.5\%} \\
	\end{tabular}
	\caption{Performance comparisons of our TwigVLM and TwigVLM++ with other token pruning methods on the \textbf{LLaVA-1.5-13B} model.}
	\label{tab:llava13b}
	\vspace{-5pt}
\end{table}

\begin{table*}[t]
\begin{center}
	\setlength{\tabcolsep}{4.pt}{
		\renewcommand{\arraystretch}{1.4}
		\footnotesize
		\centering
			\begin{tabular}{c | c c c c c c c c c | c }
				\textbf{Method} & \textbf{GQA} & \textbf{MMB} & \textbf{MME} &  \textbf{VQA}$^{\text{T}}$ & \textbf{SQA}$^{\text{I}}$ & \textbf{VQA}$^{\text{V2}}$ & \textbf{POPE} & \textbf{MMMU} & \textbf{MM-Vet} & \textbf{RelAcc}\\
				\ChangeRT{1pt}
				\rowcolor{mygray}
				\multicolumn{11}{c}{\textit{Upper Bound, 576 Tokens} \ $\textbf{(100\%)}$} \\
				\textcolor{gray}{LLaVA-1.5-7B} & \textcolor{gray}{61.9} & \textcolor{gray}{64.7} & \textcolor{gray}{1862} & \textcolor{gray}{58.2} & \textcolor{gray}{69.5} & \textcolor{gray}{78.5} & \textcolor{gray}{85.9} & \textcolor{gray}{36.3} & \textcolor{gray}{31.1} & \textcolor{gray}{100\%}  \\
				\hline
				\rowcolor{mygray}
				\multicolumn{11}{c}{\textit{Retain Averaged 192 Tokens} \ $\fg{(\downarrow 66.7\%)}$}  \\
				FastV~\cite{fastv}                   & 56.5 & 63.7 & 1786 & 57.3 & 69.5 & 74.6 & 79.2 & 35.7 & 28.1 & 95.6\% \\
				SparseVLM~\cite{sparsevlm}               & 57.6 & 62.5 & 1721 & 56.1 & 69.1 & 75.6 & 83.6 & 33.8 & 31.5 & 96.2\% \\
				PDrop~\cite{pyramiddrop}                & 57.3 & 63.3 & 1797 & 56.5 & 69.2 & 75.1 & 82.3 & - & - & 96.4\% \\
				MustDrop~\cite{mustdrop}                & 58.2 & 62.3 & 1787 & 56.5 & 69.2 & 76.0 & 82.6 & - & - & 96.6\% \\
				VisionZip~\cite{visionzip}               & 59.3 & 63.0 & 1783 & 57.3 & 68.9 & 76.8 & 85.3 & \textbf{36.6} & 31.7 & 98.5\% \\
				VisionZip\ddag~\cite{visionzip}   & 60.1 & 63.4 & 1834 & 57.8 & 68.2 & 77.4 & 84.9 & 36.2 & 32.6 & 99.2\% \\
				\textbf{TwigVLM} &  61.2 & 64.0 & 1848 & 58.0 & 68.8 & 78.1 & \textbf{87.2} & \textbf{36.6} & \textbf{32.8} & 100.3\% \\
                \textbf{TwigVLM++} & \textbf{61.2} & \textbf{64.3} & \textbf{1868} & \textbf{58.0} & \textbf{69.2} & \textbf{78.2} & 86.9 & 36.4 & 32.6 & \textbf{100.4\%} \\
				\hline
				\rowcolor{mygray}
				\multicolumn{11}{c}{\textit{Retain Averaged 128 Tokens} \ $\fg{(\downarrow 77.8\%)}$} \\
				FastV                   & 53.0 & 61.4 & 1646 & 56.0 & 69.5 & 69.2 & 73.2 & 36.3 & 28.0 & 92.1\% \\
				SparseVLM               & 56.0 & 60.0 & 1696 & 54.9 & 67.1 & 73.8 & 80.5 & 33.8 & 30.0 & 93.6\% \\
				PDrop                   & 57.1 & 61.6 & 1761 & 56.6 & 68.4 & 72.9 & 82.3 & - & - & 95.2\% \\
				MustDrop                & 56.9 & 61.1 & 1745 & 56.3 & 68.5 & 74.6 & 78.7 & - & - & 94.6\% \\
				VisionZip               & 57.6 & 62.0 & 1762 & 56.8 & 68.9 & 75.6 & 83.2 & \textbf{37.9} & 32.6 & 98.1\% \\
				VisionZip\ddag   & 58.9 & 62.6 & 1823 & 57.0 & 68.3 & 76.6 & 83.7 & 37.3 & \textbf{32.9} & 98.8\% \\
				\textbf{TwigVLM} & 60.6 & 63.5 & 1818 & 57.8 & 69.5 & 77.9 & 86.6 & 36.6 & 30.8 & 99.2\% \\
                \textbf{TwigVLM++} & \textbf{60.8} & \textbf{63.7} & \textbf{1856} & \textbf{58.0} & \textbf{69.5} & \textbf{77.9} & \textbf{87.0} & 36.4 & 31.4 & \textbf{99.8\%} \\
				\hline
				\rowcolor{mygray}
				\multicolumn{11}{c}{\textit{Retain Averaged 64 Tokens} \ $\fg{(\downarrow 88.9\%)}$}\\
				FastV                   & 44.1 & 45.9 & 1218 & 50.7 & 70.0 & 52.0 & 55.6 & 34.0 & 17.8 & 75.3\% \\
				SparseVLM               & 52.7 & 56.2 & 1505 & 51.8 & 62.2 & 68.2 & 75.1 & 32.7 & 23.3 & 85.6\% \\
				PDrop                   & 47.5 & 58.8 & 1561 & 50.6 & 69.0 & 69.2 & 55.9 & - & - & 84.4\% \\
				FasterVLM~\cite{fastervlm}              & 51.5 & 58.5 & 1573 & 53.1 & 69.6 & 66.8 & 67.2 & - & 27.5 & 87.6\% \\
				MustDrop                & 53.1 & 60.0 & 1612 & 54.2 & 63.4 & 69.3 & 68.0 & - & - & 88.1\% \\
				VisionZip               & 55.1 & 60.1 & 1690 & 55.5 & 69.0 & 72.4 & 77.0 & \textbf{36.2} & \textbf{31.7} & 94.5\% \\
				VisionZip\ddag   & 57.0 & 61.5 & 1756 & 56.0 & 68.8 & 74.2 & 80.9 & 35.6 & 30.2 & 95.6\% \\
				\textbf{TwigVLM} &  58.8 & 60.4 & 1760 & 55.8 & \textbf{70.0} & 75.6 & 82.7 & {35.9} & 29.9 & \textbf{96.3\%} \\
                \textbf{TwigVLM++} & \textbf{59.7} & \textbf{63.2} & \textbf{1801} & \textbf{56.7} & 69.5 & \textbf{76.8} & \textbf{86.6} & 35.7 & 29.0 & \textbf{97.6\%} \\
			\end{tabular}
			\caption{Performance of TwigVLM/TwigVLM++ on \textbf{LLaVA-1.5-7B} compared to existing methods under three different pruning ratios. The best result for each benchmark and pruning ratio is \textbf{bolded}.}
			\label{tab:main_more}
		}
	\end{center}
	\vspace{-5pt}
\end{table*}

\section{More Experimental Results}

\subsection{More performance comparisons}

\noindent\textbf{Comparisons on more benchmarks.} Taking LLaVA-1.5-7B as the base VLM, TABLE~\ref{tab:main_more} compares the accuracies among TwigVLM, TwigVLM++ and other visual token pruning methods on \emph{nine} VLM benchmarks under three different pruning ratios. TwigVLM and TwigVLM++ consistently outperform or match their counterparts on all benchmarks and pruning ratios, achieving the best overall RelAcc. In particular, TwigVLM and TwigVLM++ even surpass the upper bound given by the base VLM in RelAcc (100.3\%\&100.4\%) with a 66.7\% pruning ratio, demonstrating their spectacular effectiveness and robustness in accelerating VLMs to deal with various tasks.

\noindent\textbf{Comparisons on a larger base VLM.} To further demonstrate the generalization ability and superiority of our TwigVLM and TwigVLM++, we present additional experimental results on a larger VLM, LLaVA-1.5-13B, as shown in TABLE~\ref{tab:llava13b}. TwigVLM and TwigVLM++ consistently achieve the best overall RelAcc compared to all the counterparts, with their superiority being more significant as the increase of pruning ratios. These results verify the scalability and generalization ability of our TwigVLM and TwigVLM++ in accelerating large VLMs.

% To evaluate the universality of TwigVLM to VLMs beyond the LLaVA family, we also apply it to Qwen2.5-VL-7B. Since the SFT data for QwenVL is large-scale yet inaccessible, we use a 5M subset of the MAmmoTH-VL-10M dataset (single image split) \cite{mammoth-VL} as an alternative to train TwigVLM, which takes 12 hours on 8*A100 NVIDIA GPUs. The results in TABLE~\ref{tab:qwen2vl_results} show that TwigVLM can still maintain its prominent advantages over FastV in terms of accuracy and speed, showing the effectiveness of TwigVLM in practical scenarios.

\subsection{More ablation studies on TwigVLM}

\begin{table}[t]
	\vspace{-10pt}
	\centering
	\setlength{\tabcolsep}{4.0pt}
	\renewcommand{\arraystretch}{1.3}
	\small
	\begin{tabular}{l | c c  }
		\textbf{ablation variant} & ~~\textbf{TokAR} (\%)~~ & \textbf{RelSpd (\%)} \\
		\ChangeRT{1pt}
		\rowcolor{mygray}
		\multicolumn{3}{c}{\textit{Twig block initialization}\ (Table~\textcolor{iccvblue}{4c} in main text)} \\
		(a) random init.               & 37.7 & 120.4 \\
		(b) VLM layers[$L$-$T$:$L$]             & 44.1 & 131.4 \\
		(c) VLM layers[$K$:$K$+$T$]             & \textbf{57.4} & \textbf{153.6} \\
		\hline
		\rowcolor{mygray}
		\multicolumn{3}{c}{\textit{Number of twig layers}\ (Table~\textcolor{iccvblue}{4d} in main text)} \\
		(d)~$T$~=~1             & 48.7 & \textbf{154.1} \\
		(e)~$T$~=~2             & 53.4 & 152.6 \\
		(f)~$T$~=~3             & 57.4 & 153.6 \\
		(g)~$T$~=~4             & \textbf{58.1} & 145.4 \\
	\end{tabular}
	\caption{\textbf{Token acceptance rate in SSD}. We evaluate the token acceptance rate (TokAR) of the variants in the ablation experiments of the main text.} \label{tab:tar}
	\vspace{-10pt}
\end{table}

\noindent\textbf{Token acceptance rate in SSD.} In the context of speculative decoding methods \cite{specdec, liu2025kangaroo, layerskip}, the token acceptance rate (\emph{abbr.} TokAR) serves as a critical metric for assessing the efficacy of these approaches. TokAR is defined as the proportion of the draft tokens generated by the draft model that are subsequently accepted by the target model. In TwigVLM, TokAR plays a key role, which is influenced by the effectiveness of the twig block and has a significant impact on model's generation speed.

To analyze how TokAR is influenced by the design choices in TwigVLM, we evaluate this metric on several representative variants from the ablation studies presented in the main text. From the results shown in TABLE \ref{tab:tar}, we have the following findings: (i) A more effective draft model can be trained by only modifying the initialization strategy without altering the architecture. The variant (c) achieves the highest TokAR (57.4\%) and thus the highest generation speedup. (ii) Increasing the number of twig layers $T$ introduces more computational costs while improving TokAR at the same time. As a result, the RelSpd exhibits only a modest decline when $T$ increases from 1 to 3. However, it drops distinctly at $T$=4, which indicates that TokAR begins to saturate. These findings suggest that TwigVLM achieves higher speedup by striking an optimal balance between TokAR and computation costs of the draft model.

\begin{figure}
	\centering
	\begin{subfigure}{\linewidth}
        \centering
        \includegraphics[width=0.99\textwidth]{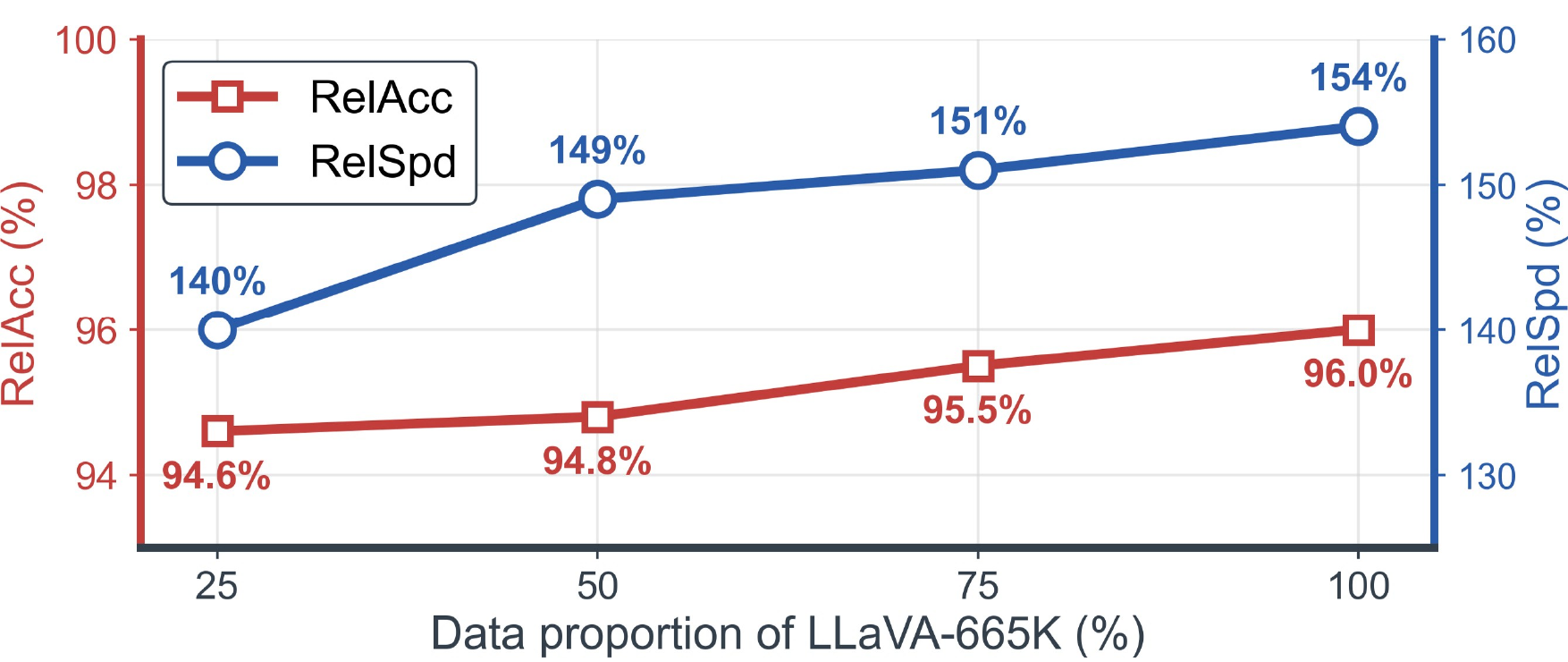}
        \subcaption{TwigVLM on LLaVA-1.5-7B}
    \end{subfigure}

    \begin{subfigure}{\linewidth}
        \centering
        \includegraphics[width=0.99\textwidth]{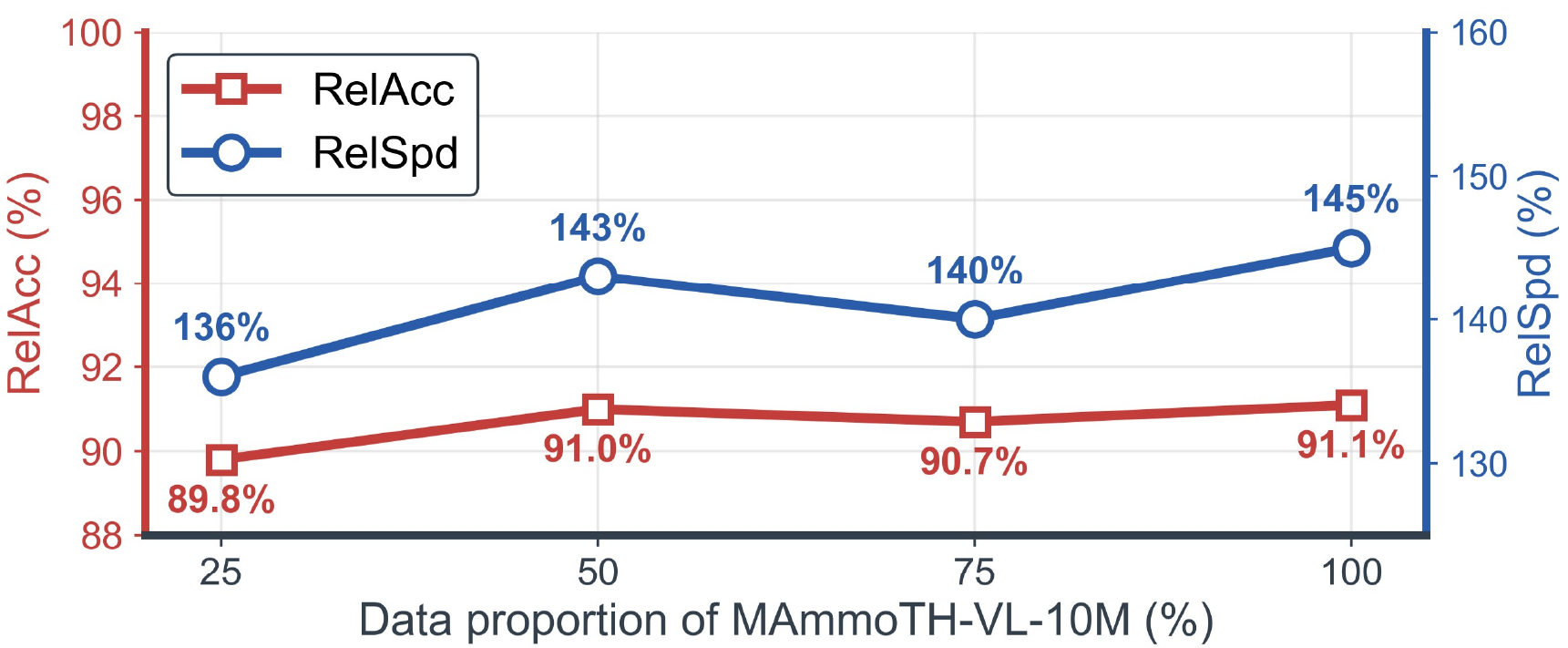}
        \subcaption{TwigVLM on Qwen2.5-VL-7B}
    \end{subfigure}
	\caption{Performance comparisons of TwigVLM models trained with \textbf{different proportions of the training dataset}. Specifically, we use LLaVA-665K to train TwigVLM models for LLaVA-1.5-7B and use MAmmoTH-VL-10M to train TwigVLM models for Qwen2.5-VL-7B. Even with only 50\% of the respective training data, TwigVLM is able to maintain competitive accuracy and speed.}
	\label{fig:data_efficiency}
	\vspace{-10pt}
\end{figure}

\noindent\textbf{Data efficiency.} To demonstrate the data efficiency of TwigVLM, we train multiple models using different proportions (i.e., 25\%, 50\%, 75\%, and 100\%) of each model's respective training dataset for LLaVA-1.5-7B and Qwen2.5-VL-7B. As shown in Fig.~\ref{fig:data_efficiency}, both models exhibit a general upward trend in accuracy and speed as the amount of training data increases. Remarkably, however, even when trained on only 50\% of their respective datasets, TwigVLM models already achieve competitive, and in some cases comparable, performance to models trained on the full dataset. Moreover, TwigVLM requires only 10\% of the training cost of the corresponding base VLM (see \ref{sec:appendix_implementation}). Recall the results in TABLE 5c of the main text, we can sum up that it is highly efficient and feasible to apply our TwigVLM and TwigVLM++ in industrial scenarios.

\noindent\textbf{Memory footprint analysis.} We measure the inference VRAM usage of the LLaVA-1.5-7B and LLaVA-Next-7B models in TABLE~\ref{tab:memory_footprint}. The introduction of the twig block brings 8\% extra VRAM cost for loading model weights. Compared to the base VLM, the overall inference VRAM cost of TwigVLM is comparable or slightly reduced due to the substantial reduction of visual tokens.
% \FloatBarrier
\begin{table}[t]
  \centering
  \vspace{-5pt}
  \renewcommand{\arraystretch}{1.2}
  \resizebox{\linewidth}{!}{%
  \begin{tabular}{lc|cc}
    model & \makecell{avg. visual\\tokens ($\bar{R}$)} & \makecell{model weights\\VRAM (GB)} & \makecell{inference\\VRAM (GB)} \\
    \ChangeRT{1pt}
    LLaVA-1.5-7B & 576 & 14.3 & 15.8 \\
    {+ TwigVLM} & 64 & 15.5 & 16.5 \\
    \hline
    LLaVA-Next-7B & 2,880 & 14.3 & 17.9 \\
    {+ TwigVLM} & 320 & 15.5 & 16.8 \\
  \end{tabular}
  }
  \caption{Memory footprint comparisons during inference.}
  \label{tab:memory_footprint}
  \vspace{-10pt}
  \end{table}

\begin{figure*}[!t]
	\begin{center}
		\begin{overpic}[width=0.92\linewidth]{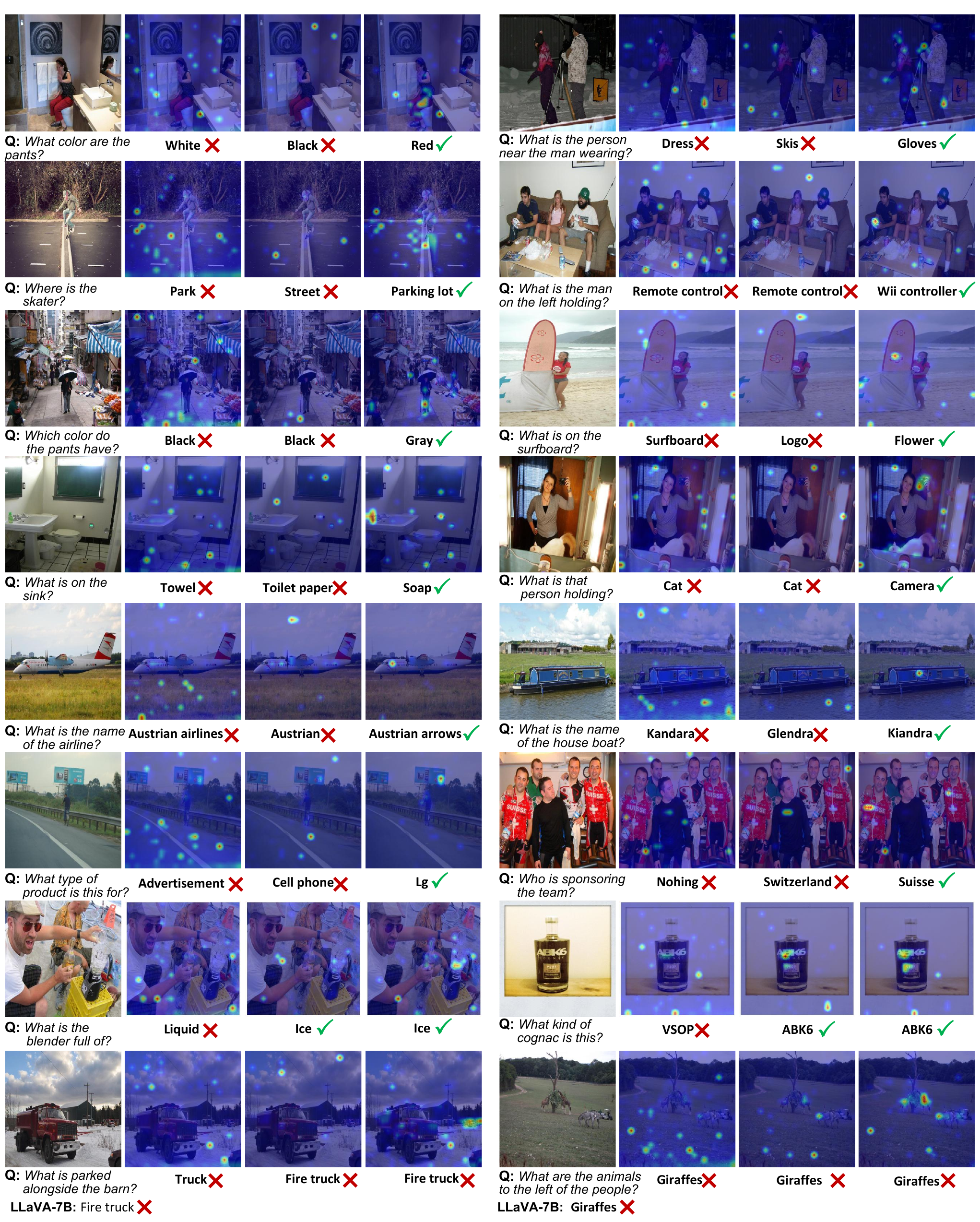}
			\put(0, 99.5){ \small {image, prompt}}
			\put(12, 99.5){ \small{FastV \cite{fastv}} }
			\put(20, 99.5){ \small {VisionZip \cite{visionzip}}}
			\put(30.5, 99.5){ \small {TwigVLM}}
			\put(40.5, 99.5){ \small {image, prompt}}
			\put(52, 99.5){ \small{FastV \cite{fastv}} }
			\put(60.5, 99.5){ \small {VisionZip \cite{visionzip}}}
			\put(71, 99.5){ \small {TwigVLM}}
		\end{overpic}
		\caption{Visualization of attention maps and predictions for FastV \cite{fastv}, VisionZip \cite{visionzip}, and our TwigVLM on the examples chosen from the GQA \cite{gqa} and TextVQA \cite{textvqa}. For the examples in the last row, we additionally provide the predictions from the LLaVA-1.5-7B.
		}\label{fig:app_fig1}
	\end{center}
\end{figure*}

\begin{figure*}[!t]
	\begin{center}
		\includegraphics[width=0.9\linewidth]{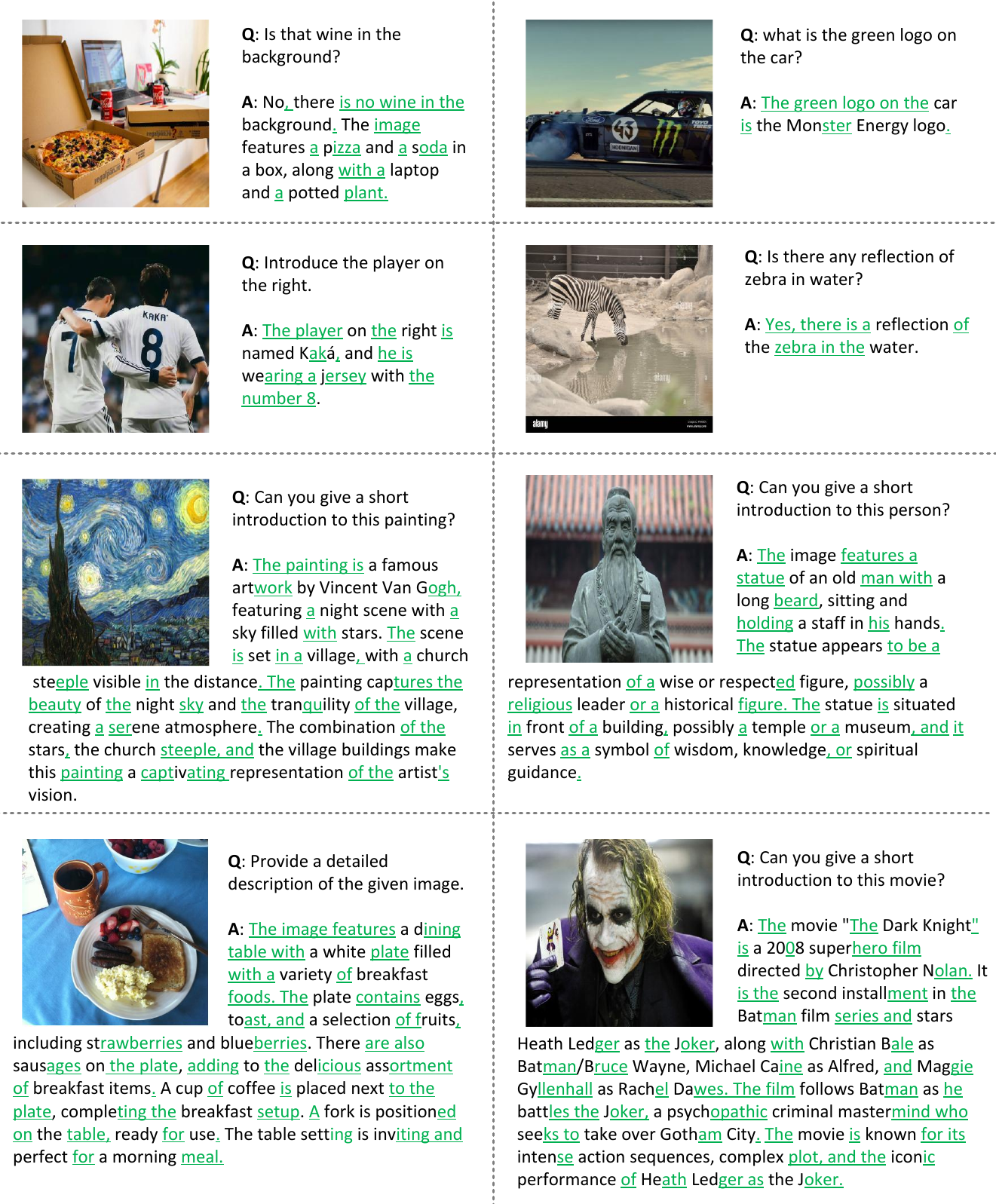}
		\caption{Examples of the generated responses using the self-speculative decoding (SSD) on MM-Vet \cite{yu2023mmvet}, with accepted tokens by the target model being highlighted in \textcolor[RGB]{0,176,80}{green}.
		}\label{fig:app_fig2}
	\end{center}
\end{figure*}

\section{More Visualized Results}

In this section, we provide more visualized results to validate the effectiveness of TwigVLM's two key components: the twig-guided visual token pruning (TTP) and self-speculative decoding (SSD). We use LLaVA-1.5-7B as the base VLM in the following experiments.

\vspace{3pt}
\noindent\textbf{Visual token pruning.} To better understand the effectiveness of the proposed TTP strategy, we compare TwigVLM with two representative token pruning methods, namely FastV \cite{fastv} and VisionZip \cite{visionzip}, by visualizing their attention map for token selection and providing the corresponding answer predictions. We provide 16 examples from the GQA and TextVQA benchmarks. As illustrated in Fig. \ref{fig:app_fig1}, TwigVLM demonstrates superior ability to comprehend the semantics in both the textual prompt and image, and accurately identify task-specific image patches (\emph{i.e.}, visual tokens), thereby activating more informative visual tokens for token pruning. In contrast, FastV and VisionZip often fail to capture the fine-grained visual details, leading to suboptimal token selection and incorrect predictions. Notably, even though TwigVLM predicts an incorrect answer, its activated visual tokens according to the attention map are reasonable. This suggests that TwigVLM's occasional failures may not be caused by the visual token pruning, but due to the limitations of the base VLM. These findings verify and explain the effectiveness of the TTP strategy.

\vspace{3pt}
\noindent\textbf{Self-speculative decoding.} To better understand the decoding behavior of the SSD strategy in TwigVLM, we show 8 examples of generated long responses on MM-Vet. From the results in Fig.~\ref{fig:app_fig2}, we have two key observations: (i) In general, the proportion of accepted tokens (in green) surpasses that of the corrected tokens (in black) by the target model, indicating that TwigVLM achieves significant speedup through its high token acceptance rate. (ii) The majority of \emph{easy} tokens, such as those associated with grammar and punctuation, are readily accepted. In contrast, the \emph{hard} tokens, which often demand complex reasoning, have a high probability of being corrected by the target model. In practice, the proportion of easy tokens is usually larger than the hard ones, which confirms the effectiveness of our SSD strategy in accelerating the decoding stage while maintaining the generation quality.

\section{Evaluation Benchmarks}
In this section, we provide a brief overview of the benchmarks used in our experiments.

\vspace{3pt}
\noindent\textbf{GQA} \cite{gqa} is a benchmark that focuses on visual scene understanding and reasoning, leveraging scene graphs, questions, and images. It incorporates spatial relationships and object properties, posing challenges for models to perform accurate visual reasoning under complex visual environments.

\vspace{3pt}
\noindent\textbf{MMBench} \cite{liu2023mmbench} adopts a hierarchical evaluation approach with three levels: Level-1 (perception and reasoning), Level-2 (six sub-abilities), and Level-3 (20 specific dimensions). This structured framework allows for a comprehensive evaluation of model performance, making it an effective tool for assessing a wide range of visual understanding capabilities. We denote it as ``MMB'' in the main text.

\vspace{3pt}
\noindent\textbf{MME} \cite{fu2023mme} assesses models across 14 subtasks that probe both perceptual and cognitive skills. Carefully crafted instruction-answer pairs guarantee a fair and comprehensive evaluation of a model's multimodal performance. The final score reported on this benchmark is the summation of both the perception and cognition scores.

\vspace{3pt}
\noindent\textbf{ScienceQA} \cite{sciqa} spans multiple scientific fields, including natural, language, and social sciences, with questions organized into 26 topics, 127 categories, and 379 skills. It evaluates a model's multimodal comprehension, multi-step reasoning, and interpretability, providing a rich testbed for assessing scientific knowledge application in visual contexts. In our experiments, we only evaluate the performance on the samples with images, denoted as ``SQA$^{\text{I}}$'' in the experimental tables.

\vspace{3pt}
\noindent\textbf{VQA-v2} \cite{vqav2} is a large-scale benchmark featuring 265K images of real-world scenes and objects, with each image paired with open-ended questions and 10 human-provided ground truth answers.

\vspace{3pt}
\noindent\textbf{TextVQA} \cite{textvqa} tests a model's ability to process and reason about text embedded within images. By requiring the integration of visual and textual information, it serves as a critical benchmark for evaluating text-based reasoning in visual contexts. To save space, we denote it as ``VQA$^{\text{T}}$'' in the experimental tables.

\vspace{3pt}
\noindent\textbf{POPE} \cite{pope} targets object hallucination evaluation by posing binary questions about object presence in images. It employs metrics such as Accuracy, Recall, Precision, and F1 score across three sampling methods. The reported score is calculated by the mean accuracy over the three indicators: adversarial, random, and popular.

\vspace{3pt}
\noindent\textbf{MMMU} \cite{mmmu} challenges models with tasks requiring college-level expertise and reasoning skills. It comprises 11.5K questions drawn from exams, quizzes, and textbooks, spanning six key disciplines: Art \& Design, Business, Science, Health \& Medicine, Humanities \& Social Science, and Tech \& Engineering. Featuring 30 subjects and 183 sub-fields, MMMU involves diverse image types, \emph{e.g.}, charts, diagrams, and chemical structures, demanding advanced perceptual and domain-specific reasoning abilities akin to those of human experts.

\vspace{3pt}
\noindent\textbf{MM-Vet} \cite{yu2023mmvet} evaluates six fundamental vision-language capabilities: recognition, OCR, knowledge, language generation, spatial awareness, and mathematical reasoning. It examines 16 specific combinations of these skills through quantitative metrics, offering a nuanced perspective on a model's proficiency in tackling intricate multimodal tasks.

\vspace{3pt}
\noindent\textbf{MMStar} \cite{mmstar} is an elite vision-indispensable multi-modal benchmark designed to rigorously evaluate the genuine multi-modal capabilities of large vision-language models. It comprises 1,500 carefully selected samples spanning six core capability dimensions, including coarse and fine-grained perception, instance reasoning, logical reasoning, science \& technology, and mathematics. Each sample is manually verified to ensure that visual content is essential for arriving at the correct answer, thereby filtering out instances answerable through text-only reasoning or dataset bias.

\vspace{3pt}
\noindent\textbf{OCRBench} \cite{ocrbench} is a comprehensive evaluation benchmark designed to assess the optical character recognition capabilities of large vision-language models across diverse text-related visual understanding tasks. It encompasses 29 sub-tasks spanning five major categories---text recognition, scene text-centric VQA, document-oriented VQA, key information extraction, and handwritten mathematical expression recognition---comprising 1,000 human-verified question-answer pairs.

\vspace{3pt}
\noindent\textbf{BLINK} \cite{blink} is a multimodal benchmark that evaluates visual perception capabilities, which are straightforward for humans yet challenging for current multimodal large language models. It reformats 14 classic computer vision tasks into 3,807 multiple-choice questions, covering relative depth estimation, jigsaw puzzle solving, visual correspondence, forensics detection, spatial reasoning, and more. The benchmark reveals that even state-of-the-art models significantly lag behind human performance on these perception-oriented tasks.

\vspace{3pt}
\noindent\textbf{Video-MME} \cite{videomme} is a comprehensive evaluation benchmark for assessing video understanding capabilities of multi-modal large language models. It comprises 900 videos totaling 254 hours with 2,700 human-annotated question-answer pairs, spanning short ($<$2 minutes), medium (4--15 minutes), and long (30--60 minutes) durations across 30 broad categories. The benchmark evaluates models in both with-subtitle and without-subtitle settings, thereby disentangling visual and textual comprehension abilities.

\vspace{3pt}
\noindent\textbf{EgoSchema} \cite{egoschema} is a diagnostic benchmark for very long-form video language understanding, featuring over 5,000 human-curated multiple-choice question-answer pairs derived from Ego4D videos. Each question requires temporal reasoning over three-minute video clips, making it substantially more demanding than prior video QA benchmarks that typically involve short clips of only a few seconds. The benchmark specifically targets the assessment of models' capacity for extended temporal comprehension and egocentric activity understanding.

\vspace{3pt}
\noindent\textbf{MVBench} \cite{mvbench} is a comprehensive benchmark designed to evaluate the temporal understanding capabilities of multi-modal video large language models. It defines 20 challenging video understanding tasks---such as action sequence recognition, scene transition detection, and attribute change identification---that specifically require dynamic, temporal reasoning rather than reliance on single-frame cues. Evaluation is conducted through a multiple-choice question-answering format, enabling scalable and reproducible assessment across diverse temporal reasoning dimensions.

\bibliographystyle{IEEEtran}
\bibliography{twigvlm}

\begin{thebibliography}{10}
\providecommand{\url}[1]{#1}
\csname url@samestyle\endcsname
\providecommand{\newblock}{\relax}
\providecommand{\bibinfo}[2]{#2}
\providecommand{\BIBentrySTDinterwordspacing}{\spaceskip=0pt\relax}
\providecommand{\BIBentryALTinterwordstretchfactor}{4}
\providecommand{\BIBentryALTinterwordspacing}{\spaceskip=\fontdimen2\font plus
\BIBentryALTinterwordstretchfactor\fontdimen3\font minus
  \fontdimen4\font\relax}
\providecommand{\BIBforeignlanguage}[2]{{%
\expandafter\ifx\csname l@#1\endcsname\relax
\typeout{** WARNING: IEEEtran.bst: No hyphenation pattern has been}%
\typeout{** loaded for the language `#1'. Using the pattern for}%
\typeout{** the default language instead.}%
\else
\language=\csname l@#1\endcsname
\fi
#2}}
\providecommand{\BIBdecl}{\relax}
\BIBdecl

\bibitem{gpt3}
T.~Brown, B.~Mann, N.~Ryder, M.~Subbiah, J.~D. Kaplan, P.~Dhariwal,
  A.~Neelakantan, P.~Shyam, G.~Sastry, A.~Askell, S.~Agarwal, A.~Herbert-Voss,
  G.~Krueger, T.~Henighan, R.~Child, A.~Ramesh, D.~Ziegler, J.~Wu, C.~Winter,
  C.~Hesse, M.~Chen, E.~Sigler, M.~Litwin, S.~Gray, B.~Chess, J.~Clark,
  C.~Berner, S.~McCandlish, A.~Radford, I.~Sutskever, and D.~Amodei, ``Language
  models are few-shot learners,'' in \emph{Advances in Neural Information
  Processing Systems}, vol.~33, 2020, pp. 1877--1901.

\bibitem{deepseekr1}
D.~Guo, D.~Yang, H.~Zhang, J.~Song, R.~Zhang, R.~Xu, Q.~Zhu, S.~Ma, P.~Wang,
  X.~Bi \emph{et~al.}, ``Deepseek-r1: Incentivizing reasoning capability in
  llms via reinforcement learning,'' \emph{arXiv preprint arXiv:2501.12948},
  2025.

\bibitem{gpt4v}
OpenAI, ``Gpt-4v(ision) system card,'' OpenAI, Tech. Rep., 2023.

\bibitem{llava1}
H.~Liu, C.~Li, Q.~Wu, and Y.~J. Lee, ``Visual instruction tuning,'' in
  \emph{Advances in Neural Information Processing Systems}, vol.~36, 2023, pp.
  34\,892--34\,916.

\bibitem{qwenvl}
J.~Bai, S.~Bai, S.~Yang, S.~Wang, S.~Tan, P.~Wang, J.~Lin, C.~Zhou, and
  J.~Zhou, ``Qwen-vl: A frontier large vision-language model with versatile
  abilities,'' \emph{arXiv preprint arXiv:2308.12966}, 2023.

\bibitem{internvl}
Z.~Chen, W.~Wang, H.~Tian, S.~Ye, Z.~Gao, E.~Cui, W.~Tong, K.~Hu, J.~Luo, Z.~Ma
  \emph{et~al.}, ``How far are we to gpt-4v? closing the gap to commercial
  multimodal models with open-source suites,'' \emph{Science China Information
  Sciences}, vol.~67, no.~12, p. 220101, 2024.

\bibitem{shao2025imp}
Z.~Shao, Z.~Yu, J.~Yu, X.~Ouyang, L.~Zheng, Z.~Gai, M.~Wang, Z.~Kuang, and
  J.~Ding, ``Imp: Highly capable large multimodal models for mobile devices,''
  \emph{IEEE Transactions on Multimedia}, 2025.

\bibitem{peng2024grounding}
Z.~Peng, W.~Wang, L.~Dong, Y.~Hao, S.~Huang, S.~Ma, Q.~Ye, and F.~Wei,
  ``Grounding multimodal large language models to the world,'' in \emph{The
  Twelfth International Conference on Learning Representations}, 2024.

\bibitem{ma2024groma}
C.~Ma, Y.~Jiang, J.~Wu, Z.~Yuan, and X.~Qi, ``Groma: Localized visual
  tokenization for grounding multimodal large language models,'' in
  \emph{European Conference on Computer Vision}.\hskip 1em plus 0.5em minus
  0.4em\relax Springer, 2024, pp. 417--435.

\bibitem{ye2023mplug}
J.~Ye, A.~Hu, H.~Xu, Q.~Ye, M.~Yan, Y.~Dan, C.~Zhao, G.~Xu, C.~Li, J.~Tian
  \emph{et~al.}, ``mplug-docowl: Modularized multimodal large language model
  for document understanding,'' \emph{arXiv preprint arXiv:2307.02499}, 2023.

\bibitem{luo2024layoutllm}
C.~Luo, Y.~Shen, Z.~Zhu, Q.~Zheng, Z.~Yu, and C.~Yao, ``Layoutllm: Layout
  instruction tuning with large language models for document understanding,''
  in \emph{Proceedings of the IEEE/CVF conference on computer vision and
  pattern recognition}, 2024, pp. 15\,630--15\,640.

\bibitem{wang2024mobile}
J.~Wang, H.~Xu, J.~Ye, M.~Yan, W.~Shen, J.~Zhang, F.~Huang, and J.~Sang,
  ``Mobile-agent: Autonomous multi-modal mobile device agent with visual
  perception,'' \emph{arXiv preprint arXiv:2401.16158}, 2024.

\bibitem{zhang2023appagent}
C.~Zhang, Z.~Yang, J.~Liu, Y.~Han, X.~Chen, Z.~Huang, B.~Fu, and G.~Yu,
  ``Appagent: Multimodal agents as smartphone users,'' \emph{arXiv preprint
  arXiv:2312.13771}, 2023.

\bibitem{brohan2023rt}
A.~Brohan, N.~Brown, J.~Carbajal, Y.~Chebotar, X.~Chen, K.~Choromanski,
  T.~Ding, D.~Driess, A.~Dubey, C.~Finn \emph{et~al.}, ``Rt-2:
  Vision-language-action models transfer web knowledge to robotic control,''
  \emph{arXiv preprint arXiv:2307.15818}, 2023.

\bibitem{kim2024openvla}
M.~J. Kim, K.~Pertsch, S.~Karamcheti, T.~Xiao, A.~Balakrishna, S.~Nair,
  R.~Rafailov, E.~Foster, G.~Lam, P.~Sanketi \emph{et~al.}, ``Openvla: An
  open-source vision-language-action model,'' \emph{arXiv preprint
  arXiv:2406.09246}, 2024.

\bibitem{transformer}
A.~Vaswani, N.~Shazeer, N.~Parmar, J.~Uszkoreit, L.~Jones, A.~N. Gomez, L.~u.
  Kaiser, and I.~Polosukhin, ``Attention is all you need,'' in \emph{Advances
  in Neural Information Processing Systems}, vol.~30, 2017.

\bibitem{fastv}
L.~Chen, H.~Zhao, T.~Liu, S.~Bai, J.~Lin, C.~Zhou, and B.~Chang, ``An image is
  worth 1/2 tokens after layer 2: Plug-and-play inference acceleration for
  large vision-language models,'' in \emph{European Conference on Computer
  Vision}.\hskip 1em plus 0.5em minus 0.4em\relax Springer, 2024, pp. 19--35.

\bibitem{sparsevlm}
Y.~Zhang, C.-K. Fan, J.~Ma, W.~Zheng, T.~Huang, K.~Cheng, D.~Gudovskiy,
  T.~Okuno, Y.~Nakata, K.~Keutzer \emph{et~al.}, ``Sparsevlm: Visual token
  sparsification for efficient vision-language model inference,'' \emph{arXiv
  preprint arXiv:2410.04417}, 2024.

\bibitem{mustdrop}
T.~Liu, L.~Shi, R.~Hong, Y.~Hu, Q.~Yin, and L.~Zhang, ``Multi-stage vision
  token dropping: Towards efficient multimodal large language model,''
  \emph{arXiv preprint arXiv:2411.10803}, 2024.

\bibitem{visionzip}
S.~Yang, Y.~Chen, Z.~Tian, C.~Wang, J.~Li, B.~Yu, and J.~Jia, ``Visionzip:
  Longer is better but not necessary in vision language models,'' \emph{arXiv
  preprint arXiv:2412.04467}, 2024.

\bibitem{fastervlm}
Q.~Zhang, A.~Cheng, M.~Lu, Z.~Zhuo, M.~Wang, J.~Cao, S.~Guo, Q.~She, and
  S.~Zhang, ``[cls] attention is all you need for training-free visual token
  pruning: Make vlm inference faster,'' \emph{arXiv preprint arXiv:2412.01818},
  2024.

\bibitem{layerskip}
M.~Elhoushi, A.~Shrivastava, D.~Liskovich, B.~Hosmer, B.~Wasti, L.~Lai,
  A.~Mahmoud, B.~Acun, S.~Agarwal, A.~Roman \emph{et~al.}, ``Layerskip:
  Enabling early exit inference and self-speculative decoding,'' \emph{arXiv
  preprint arXiv:2404.16710}, 2024.

\bibitem{llava1.5}
H.~Liu, C.~Li, Y.~Li, and Y.~J. Lee, ``Improved baselines with visual
  instruction tuning,'' in \emph{Proceedings of the IEEE/CVF Conference on
  Computer Vision and Pattern Recognition}, 2024, pp. 26\,296--26\,306.

\bibitem{twigvlm_v1}
Z.~Shao, M.~Wang, Z.~Yu, W.~Pan, Y.~Yang, T.~Wei, H.~Zhang, N.~Mao, W.~Chen,
  and J.~Yu, ``Growing a twig to accelerate large vision-language models,'' in
  \emph{Proceedings of the IEEE/CVF International Conference on Computer Vision
  (ICCV)}, 2025, pp. 20\,064--20\,074.

\bibitem{mamba}
A.~Gu and T.~Dao, ``Mamba: Linear-time sequence modeling with selective state
  spaces,'' in \emph{First Conference on Language Modeling}, 2024.

\bibitem{yoco}
Y.~Sun, L.~Dong, Y.~Zhu, S.~Huang, W.~Wang, S.~Ma, Q.~Zhang, J.~Wang, and
  F.~Wei, ``You only cache once: Decoder-decoder architectures for language
  models,'' \emph{Advances in Neural Information Processing Systems}, vol.~37,
  pp. 7339--7361, 2025.

\bibitem{deepseekv3}
A.~Liu, B.~Feng, B.~Xue, B.~Wang, B.~Wu, C.~Lu, C.~Zhao, C.~Deng, C.~Zhang,
  C.~Ruan \emph{et~al.}, ``Deepseek-v3 technical report,'' \emph{arXiv preprint
  arXiv:2412.19437}, 2024.

\bibitem{llmlingua}
H.~Jiang, Q.~Wu, C.-Y. Lin, Y.~Yang, and L.~Qiu, ``Llmlingua: Compressing
  prompts for accelerated inference of large language models,'' in
  \emph{Proceedings of the 2023 Conference on Empirical Methods in Natural
  Language Processing}, 2023, pp. 13\,358--13\,376.

\bibitem{gist}
J.~Mu, X.~Li, and N.~Goodman, ``Learning to compress prompts with gist
  tokens,'' \emph{Advances in Neural Information Processing Systems}, vol.~36,
  pp. 19\,327--19\,352, 2023.

\bibitem{flashattention}
T.~Dao, D.~Y. Fu, S.~Ermon, A.~Rudra, and C.~R{\'e}, ``Flash{A}ttention: Fast
  and memory-efficient exact attention with {IO}-awareness,'' in \emph{Advances
  in Neural Information Processing Systems (NeurIPS)}, 2022.

\bibitem{sparseattn}
S.~Dai, H.~Genc, R.~Venkatesan, and B.~Khailany, ``Efficient transformer
  inference with statically structured sparse attention,'' in \emph{2023 60th
  ACM/IEEE Design Automation Conference (DAC)}.\hskip 1em plus 0.5em minus
  0.4em\relax IEEE, 2023, pp. 1--6.

\bibitem{streamingllm}
G.~Xiao, Y.~Tian, B.~Chen, S.~Han, and M.~Lewis, ``Efficient streaming language
  models with attention sinks,'' in \emph{The Twelfth International Conference
  on Learning Representations}, 2024.

\bibitem{h2o}
Z.~Zhang, Y.~Sheng, T.~Zhou, T.~Chen, L.~Zheng, R.~Cai, Z.~Song, Y.~Tian,
  C.~R{\'e}, C.~Barrett \emph{et~al.}, ``H2o: Heavy-hitter oracle for efficient
  generative inference of large language models,'' \emph{Advances in Neural
  Information Processing Systems}, vol.~36, pp. 34\,661--34\,710, 2023.

\bibitem{specdec}
Y.~Leviathan, M.~Kalman, and Y.~Matias, ``Fast inference from transformers via
  speculative decoding,'' in \emph{International Conference on Machine
  Learning}.\hskip 1em plus 0.5em minus 0.4em\relax PMLR, 2023, pp.
  19\,274--19\,286.

\bibitem{medusa}
T.~Cai, Y.~Li, Z.~Geng, H.~Peng, J.~D. Lee, D.~Chen, and T.~Dao, ``Medusa:
  Simple llm inference acceleration framework with multiple decoding heads,''
  \emph{arXiv preprint arXiv:2401.10774}, 2024.

\bibitem{liu2025kangaroo}
F.~Liu, Y.~Tang, Z.~Liu, Y.~Ni, D.~Tang, K.~Han, and Y.~Wang, ``Kangaroo:
  Lossless self-speculative decoding for accelerating llms via double early
  exiting,'' \emph{Advances in Neural Information Processing Systems}, vol.~37,
  pp. 11\,946--11\,965, 2025.

\bibitem{zhang2024cross}
Z.~Zhang, S.~Yadav, F.~Han, and E.~Shutova, ``Cross-modal information flow in
  multimodal large language models,'' \emph{arXiv preprint arXiv:2411.18620},
  2024.

\bibitem{ye2024fit}
W.~Ye, Q.~Wu, W.~Lin, and Y.~Zhou, ``Fit and prune: Fast and training-free
  visual token pruning for multi-modal large language models,'' \emph{arXiv
  preprint arXiv:2409.10197}, 2024.

\bibitem{ye2024atp}
X.~Ye, Y.~Gan, Y.~Ge, X.-P. Zhang, and Y.~Tang, ``Atp-llava: Adaptive token
  pruning for large vision language models,'' \emph{arXiv preprint
  arXiv:2412.00447}, 2024.

\bibitem{dyvte}
Q.~Wu, W.~Lin, W.~Ye, Y.~Zhou, X.~Sun, and R.~Ji, ``Accelerating multimodal
  large language models via dynamic visual-token exit and the empirical
  findings,'' \emph{arXiv preprint arXiv:2411.19628}, 2024.

\bibitem{FiCoCo-V}
Y.~Han, X.~Liu, P.~Ding, D.~Wang, H.~Chen, Q.~Yan, and S.~Huang, ``Rethinking
  token reduction in mllms: Towards a unified paradigm for training-free
  acceleration,'' \emph{arXiv preprint arXiv:2411.17686}, 2024.

\bibitem{llavolta}
J.~Chen, L.~Ye, J.~He, Z.-Y. Wang, D.~Khashabi, and A.~Yuille, ``Llavolta:
  Efficient multi-modal models via stage-wise visual context compression,''
  \emph{arXiv preprint arXiv:2406.20092}, 2024.

\bibitem{pyramiddrop}
L.~Xing, Q.~Huang, X.~Dong, J.~Lu, P.~Zhang, Y.~Zang, Y.~Cao, C.~He, J.~Wang,
  F.~Wu \emph{et~al.}, ``Pyramiddrop: Accelerating your large vision-language
  models via pyramid visual redundancy reduction,'' \emph{arXiv preprint
  arXiv:2410.17247}, 2024.

\bibitem{yopo}
Z.~Zhang, P.~Pham, W.~Zhao, K.~Wan, Y.-J. Li, J.~Zhou, D.~Miranda, A.~Kale, and
  C.~Xu, ``Treat visual tokens as text? but your mllm only needs fewer efforts
  to see,'' \emph{arXiv preprint arXiv:2410.06169}, 2024.

\bibitem{chai2024auroracap}
W.~Chai, E.~Song, Y.~Du, C.~Meng, V.~Madhavan, O.~Bar-Tal, J.-N. Hwang, S.~Xie,
  and C.~D. Manning, ``Auroracap: Efficient, performant video detailed
  captioning and a new benchmark,'' \emph{arXiv preprint arXiv:2410.03051},
  2024.

\bibitem{tome}
D.~Bolya, C.-Y. Fu, X.~Dai, P.~Zhang, C.~Feichtenhofer, and J.~Hoffman, ``Token
  merging: Your vit but faster,'' in \emph{The Eleventh International
  Conference on Learning Representations}, 2023.

\bibitem{prumerge}
Y.~Shang, M.~Cai, B.~Xu, Y.~J. Lee, and Y.~Yan, ``Llava-prumerge: Adaptive
  token reduction for efficient large multimodal models,'' \emph{arXiv preprint
  arXiv:2403.15388}, 2024.

\bibitem{fastvlm}
P.~K.~A. Vasu, F.~Faghri, C.-L. Li, C.~Koc, N.~True, A.~Antony, G.~Santhanam,
  J.~Gabriel, P.~Grasch, O.~Tuzel \emph{et~al.}, ``Fastvlm: Efficient vision
  encoding for vision language models,'' \emph{arXiv preprint
  arXiv:2412.13303}, 2024.

\bibitem{zhang2025beyond}
Q.~Zhang, A.~Cheng, M.~Lu, R.~Zhang, Z.~Zhuo, J.~Cao, S.~Guo, Q.~She, and
  S.~Zhang, ``Beyond text-visual attention: Exploiting visual cues for
  effective token pruning in vlms,'' in \emph{Proceedings of the IEEE/CVF
  International Conference on Computer Vision}, 2025, pp. 20\,857--20\,867.

\bibitem{jiang2025what}
Y.~Jiang, Q.~Wu, W.~Lin, W.~Yu, and Y.~Zhou, ``What kind of visual tokens do we
  need? training-free visual token pruning for multi-modal large language
  models from the perspective of graph,'' in \emph{Proceedings of the AAAI
  Conference on Artificial Intelligence}, 2025, aAAI 2025.

\bibitem{endo2025feather}
M.~Endo, X.~Wang, and S.~Yeung-Levy, ``Feather the throttle: Revisiting visual
  token pruning for vision-language model acceleration,'' in \emph{Proceedings
  of the IEEE/CVF International Conference on Computer Vision}, 2025, pp.
  22\,826--22\,835.

\bibitem{wen2025token}
Z.~Wen, Y.~Gao, W.~Li, C.~He, and L.~Zhang, ``Token pruning in multimodal large
  language models: Are we solving the right problem?'' in \emph{Findings of the
  Association for Computational Linguistics: ACL 2025}, 2025, pp.
  15\,537--15\,549.

\bibitem{gagrani2024speculative}
M.~Gagrani, R.~Goel, W.~Jeon, J.~Park, M.~Lee, and C.~Lott, ``On speculative
  decoding for multimodal large language models,'' in \emph{Proceedings of the
  IEEE/CVF Conference on Computer Vision and Pattern Recognition}, 2024, pp.
  8285--8289.

\bibitem{sgl}
W.~Zhao, Y.~Han, J.~Tang, Z.~Li, Y.~Song, K.~Wang, Z.~Wang, and Y.~You, ``A
  stitch in time saves nine: Small vlm is a precise guidance for accelerating
  large vlms,'' \emph{arXiv preprint arXiv:2412.03324}, 2024.

\bibitem{huo2025specllava}
\BIBentryALTinterwordspacing
M.~Huo, J.~Zhang, H.~Wang, J.~Xu, Z.~Chen, H.~Tai, and Y.~Chen, ``Spec-llava:
  Accelerating vision-language models with dynamic tree-based speculative
  decoding,'' 2025. [Online]. Available: \url{https://arxiv.org/abs/2509.11961}
\BIBentrySTDinterwordspacing

\bibitem{huang2025specvlmfast}
\BIBentryALTinterwordspacing
H.~Huang, F.~Yang, Z.~Liu, X.~Yin, D.~Li, P.~Ren, and E.~Barsoum, ``Specvlm:
  Fast speculative decoding in vision-language models,'' \emph{CoRR}, vol.
  abs/2509.11815, 2025. [Online]. Available:
  \url{https://arxiv.org/abs/2509.11815}
\BIBentrySTDinterwordspacing

\bibitem{ji-etal-2025-specvlm}
\BIBentryALTinterwordspacing
Y.~Ji, J.~Zhang, H.~Xia, J.~Chen, L.~Shou, G.~Chen, and H.~Li, ``{S}pec{VLM}:
  Enhancing speculative decoding of video {LLM}s via verifier-guided token
  pruning,'' in \emph{Proceedings of the 2025 Conference on Empirical Methods
  in Natural Language Processing}, C.~Christodoulopoulos, T.~Chakraborty,
  C.~Rose, and V.~Peng, Eds.\hskip 1em plus 0.5em minus 0.4em\relax Suzhou,
  China: Association for Computational Linguistics, Nov. 2025, pp. 7205--7219.
  [Online]. Available: \url{https://aclanthology.org/2025.emnlp-main.366/}
\BIBentrySTDinterwordspacing

\bibitem{kang2025vispec}
\BIBentryALTinterwordspacing
J.~Kang, H.~Shu, W.~Li, Y.~Zhai, and X.~Chen, ``Vispec: Accelerating
  vision-language models with vision-aware speculative decoding,'' in
  \emph{Advances in Neural Information Processing Systems 38 (NeurIPS 2025)},
  2025. [Online]. Available: \url{https://openreview.net/forum?id=x2BsIdJJJW}
\BIBentrySTDinterwordspacing

\bibitem{zhou2024survey}
Z.~Zhou, X.~Ning, K.~Hong, T.~Fu, J.~Xu, S.~Li, Y.~Lou, L.~Wang, Z.~Yuan, X.~Li
  \emph{et~al.}, ``A survey on efficient inference for large language models,''
  \emph{arXiv preprint arXiv:2404.14294}, 2024.

\bibitem{gqa}
D.~A. Hudson and C.~D. Manning, ``Gqa: A new dataset for real-world visual
  reasoning and compositional question answering,'' in \emph{Proceedings of the
  IEEE/CVF Conference on Computer Vision and Pattern Recognition}, 2019.

\bibitem{textvqa}
A.~Singh, V.~Natarajan, M.~Shah, Y.~Jiang, X.~Chen, D.~Batra, D.~Parikh, and
  M.~Rohrbach, ``Towards vqa models that can read,'' in \emph{Proceedings of
  the IEEE/CVF Conference on Computer Vision and Pattern Recognition}, 2019,
  pp. 8317--8326.

\bibitem{jepo}
\BIBentryALTinterwordspacing
Y.~Tang, S.~Wang, L.~Madaan, and R.~Munos, ``Beyond verifiable rewards: Scaling
  reinforcement learning in language models to unverifiable data,'' in
  \emph{Advances in Neural Information Processing Systems}, 2025. [Online].
  Available: \url{https://openreview.net/forum?id=pc6M9h3T9m}
\BIBentrySTDinterwordspacing

\bibitem{verifree}
\BIBentryALTinterwordspacing
X.~Zhou, Z.~Liu, A.~Sims, H.~Wang, T.~Pang, C.~Li, L.~Wang, M.~Lin, and C.~Du,
  ``Reinforcing general reasoning without verifiers,'' in \emph{International
  Conference on Learning Representations}, 2026. [Online]. Available:
  \url{https://openreview.net/forum?id=nnwvwge40d}
\BIBentrySTDinterwordspacing

\bibitem{grpo}
Z.~Shao, P.~Wang, Q.~Zhu, R.~Xu, J.~Song, M.~Zhang, Y.~K. Li, Y.~Wu, and
  D.~Guo, ``Deepseekmath: Pushing the limits of mathematical reasoning in open
  language models,'' \emph{arXiv preprint arXiv:2402.03300}, 2024.

\bibitem{specinfer}
X.~Miao, G.~Oliaro, Z.~Zhang, X.~Cheng, Z.~Wang, Z.~Zhang, R.~Y.~Y. Wong,
  A.~Zhu, L.~Yang, X.~Shi, C.~Shi, Z.~Chen, D.~Arfeen, R.~Cen, and Z.~Jia,
  ``Specinfer: Accelerating large language model serving with tree-based
  speculative inference and verification,'' in \emph{Proceedings of the 29th
  ACM International Conference on Architectural Support for Programming
  Languages and Operating Systems (ASPLOS)}, 2024.

\bibitem{llavanext}
\BIBentryALTinterwordspacing
H.~Liu, C.~Li, Y.~Li, B.~Li, Y.~Zhang, S.~Shen, and Y.~J. Lee, ``Llava-next:
  Improved reasoning, ocr, and world knowledge,'' January 2024. [Online].
  Available: \url{https://llava-vl.github.io/blog/2024-01-30-llava-next/}
\BIBentrySTDinterwordspacing

\bibitem{mammoth-VL}
\BIBentryALTinterwordspacing
J.~Guo, T.~Zheng, Y.~Bai, B.~Li, Y.~Wang, K.~Zhu, Y.~Li, G.~Neubig, W.~Chen,
  and X.~Yue, ``Mammoth-vl: Eliciting multimodal reasoning with instruction
  tuning at scale,'' 2024. [Online]. Available:
  \url{https://arxiv.org/abs/2412.05237}
\BIBentrySTDinterwordspacing

\bibitem{liu2023mmbench}
Y.~Liu, H.~Duan, Y.~Zhang, B.~Li, S.~Zhang, W.~Zhao, Y.~Yuan, J.~Wang, C.~He,
  Z.~Liu \emph{et~al.}, ``Mmbench: Is your multi-modal model an all-around
  player?'' in \emph{European Conference on Computer Vision}.\hskip 1em plus
  0.5em minus 0.4em\relax Springer, 2025, pp. 216--233.

\bibitem{fu2023mme}
C.~Fu, P.~Chen, Y.~Shen, Y.~Qin, M.~Zhang, X.~Lin, Z.~Qiu, W.~Lin, J.~Yang,
  X.~Zheng \emph{et~al.}, ``Mme: A comprehensive evaluation benchmark for
  multimodal large language models,'' \emph{arXiv preprint arXiv:2306.13394},
  2023.

\bibitem{sciqa}
P.~Lu, S.~Mishra, T.~Xia, L.~Qiu, K.-W. Chang, S.-C. Zhu, O.~Tafjord, P.~Clark,
  and A.~Kalyan, ``Learn to explain: Multimodal reasoning via thought chains
  for science question answering,'' \emph{Advances in Neural Information
  Processing Systems}, vol.~35, pp. 2507--2521, 2022.

\bibitem{vqav2}
Y.~Goyal, T.~Khot, D.~Summers-Stay, D.~Batra, and D.~Parikh, ``Making the v in
  vqa matter: Elevating the role of image understanding in visual question
  answering,'' in \emph{Proceedings of the IEEE/CVF Conference on Computer
  Vision and Pattern Recognition}, 2017, pp. 6904--6913.

\bibitem{yu2023mmvet}
W.~Yu, Z.~Yang, L.~Li, J.~Wang, K.~Lin, Z.~Liu, X.~Wang, and L.~Wang, ``Mm-vet:
  Evaluating large multimodal models for integrated capabilities,'' \emph{arXiv
  preprint arXiv:2308.02490}, 2023.

\bibitem{pope}
Y.~Li, Y.~Du, K.~Zhou, J.~Wang, W.~X. Zhao, and J.-R. Wen, ``Evaluating object
  hallucination in large vision-language models,'' \emph{arXiv:2305.10355},
  2023.

\bibitem{mmmu}
X.~Yue, Y.~Ni, K.~Zhang, T.~Zheng, R.~Liu, G.~Zhang, S.~Stevens, D.~Jiang,
  W.~Ren, Y.~Sun \emph{et~al.}, ``Mmmu: A massive multi-discipline multimodal
  understanding and reasoning benchmark for expert agi,'' in \emph{Proceedings
  of the IEEE/CVF Conference on Computer Vision and Pattern Recognition}, 2024,
  pp. 9556--9567.

\bibitem{mmstar}
L.~Chen, J.~Li, X.~Dong, P.~Zhang, Y.~Zang, Z.~Chen, H.~Duan, J.~Wang, Y.~Qiao,
  D.~Lin, and F.~Zhao, ``Are we on the right way for evaluating large
  vision-language models?'' in \emph{Advances in Neural Information Processing
  Systems}, 2024.

\bibitem{ocrbench}
Y.~Liu, Z.~Li, M.~Huang, B.~Yang, W.~Yu, C.~Li, X.-C. Yin, C.-L. Liu, L.~Jin,
  and X.~Bai, ``Ocrbench: On the hidden mystery of ocr in large multimodal
  models,'' \emph{Science China Information Sciences}, vol.~67, no.~12, 2024.

\bibitem{blink}
X.~Fu, Y.~Hu, B.~Li, Y.~Feng, H.~Wang, X.~Lin, D.~Roth, N.~A. Smith, W.-C. Ma,
  and R.~Krishna, ``Blink: Multimodal large language models can see but not
  perceive,'' in \emph{Proceedings of the European Conference on Computer
  Vision}, 2024.

\bibitem{videomme}
C.~Fu, Y.~Dai, Y.~Luo, L.~Li, S.~Ren, R.~Zhang, Z.~Wang, C.~Zhou, Y.~Shen,
  M.~Zhang \emph{et~al.}, ``Video-mme: The first-ever comprehensive evaluation
  benchmark of multi-modal llms in video analysis,'' in \emph{Proceedings of
  the IEEE/CVF Conference on Computer Vision and Pattern Recognition}, 2025.

\bibitem{egoschema}
K.~Mangalam, R.~Akshulakov, and J.~Malik, ``Egoschema: A diagnostic benchmark
  for very long-form video language understanding,'' in \emph{Advances in
  Neural Information Processing Systems}, 2023.

\bibitem{mvbench}
K.~Li, Y.~Wang, Y.~He, Y.~Li, Y.~Wang, Y.~Liu, Z.~Wang, J.~Xu, G.~Chen, P.~Lou,
  L.~Wang, and Y.~Qiao, ``Mvbench: A comprehensive multi-modal video
  understanding benchmark,'' in \emph{Proceedings of the IEEE/CVF Conference on
  Computer Vision and Pattern Recognition}, 2024, pp. 22\,195--22\,206.

\end{thebibliography}

\end{document}